\documentclass[1p, times]{elsarticle}

\RequirePackage{amsthm,amsmath,amsfonts,amssymb}
\usepackage{graphicx}
\usepackage{enumerate}
\usepackage{natbib}
\usepackage[utf8]{inputenc}
\usepackage{xcolor}
\usepackage{natbib}
\usepackage[hyphens]{url}
\usepackage{mathrsfs}
\usepackage{amsfonts}
\usepackage{amsthm}
\usepackage{comment}
\usepackage{arydshln}
\usepackage{multirow}
\usepackage{subcaption}
\usepackage{tikz}
\usetikzlibrary{arrows}
\usetikzlibrary{positioning}

\usepackage{setspace}
\theoremstyle{plain}
\newtheorem{property}{Property}

\begin{document}
\begin{frontmatter}
\title{Indiscriminate Disruption of Conditional Inference on Multivariate Gaussians}

\author[afit]{William N. Caballero}
\ead{william.caballero@afit.edu}
\author[duke2]{Matthew LaRosa}
\ead{matthew.larosa@duke.edu}
\author[duke1]{Alexander Fisher}
\ead{alexander.fisher@duke.edu}
\author[duke2]{Vahid Tarokh}
\ead{vahid.tarokh@duke.edu}
\affiliation[afit]{organization={Department of Operational Sciences, Air Force Institute of Technology},
             addressline={2950 Hobson Way},
             city={Wright Patterson AFB},
             postcode={45433},
             state={OH},
             country={USA}
             }
\affiliation[duke2]{organization={Department of Electrical and Computer Engineering, Duke University},
             addressline={130 Hudson Hall},
             city={Durham},
             postcode={27708},
             state={NC},
             country={USA}
             }
\affiliation[duke1]{organization={Department of Statistical Science, Duke University},
             addressline={214 Old Chemistry},
             city={Durham},
             postcode={27708},
             state={NC},
             country={USA}
             }

\begin{abstract}
The multivariate Gaussian distribution underpins myriad operations-research, decision-analytic, and machine-learning models (e.g., Bayesian optimization, Gaussian influence diagrams, and variational autoencoders). However, despite recent advances in adversarial machine learning (AML), inference for Gaussian models in the presence of an adversary is notably understudied. Therefore, we consider a self-interested attacker who wishes to disrupt a decisionmaker's conditional inference and subsequent actions by corrupting a set of evidentiary variables. To avoid detection, the attacker also desires the attack to appear plausible wherein plausibility is determined by the density of the corrupted evidence. We consider white- and grey-box settings such that the attacker has complete and incomplete knowledge about the decisionmaker’s underlying multivariate Gaussian distribution, respectively. Select instances are shown to reduce to quadratic and stochastic quadratic programs, and structural properties are derived to inform solution methods. We assess the impact and efficacy of these attacks in three examples, including, real estate evaluation, interest rate estimation and signals processing. Each example leverages an alternative underlying model, thereby highlighting the attacks' broad applicability. Through these applications, we also juxtapose the behavior of the white- and grey-box attacks to understand how uncertainty and structure affect attacker behavior. 
\end{abstract}




\end{frontmatter}


\section{Introduction}

Multivariate Gaussians (MVGs) underpin a host of probabilistic models used in decision making tasks, especially popular machine learning procedures \citep{bishop2006pattern}.
From image classification via Gaussian discriminant analysis \citep{hastie1996discriminant, das2021method} and MVG descriptors \citep{grana2013image} to
satellite attitude estimation via the regular Kalman filter \citep{soken2014robust},
inference and uncertainty quantification under MVGs is well studied in a variety of contexts \citep{roweis1999unifying}.
Increasingly, however, fatal weaknesses are found when automated decision making processes confront an adversarial opponent \citep{BIGGIO}.
For example, \cite{szegedy2013intriguing} find that nearly imperceptible perturbation of input data can achieve robust misclassification on neural networks built from different training data. Indeed, \cite{gu2017badnets} strategically train a street sign classifier to behave as expected in a majority of scenarios, yet the presence of a small sticker on a traffic stop sign causes the object to be inappropriately classified as a speed-limit sign. Despite the illustrated success of AML and the ubiquity of MVGs, tailored MVGs attacks are not well-studied.

Nevertheless, the efficacy of the aforementioned attacks, coupled with their demonstrable covertness, makes it the duty of a decisionmaker to design automated decisionmaking procedures that proactively guard against adversarial attacks.
\cite{rios2023adversarial} explicitly model how a decisionmaker may update beliefs about the nature of an attacker iteratively within a Bayesian framework. However, this approach hinges upon a detailed understanding of the attacker's problem and their ability to manipulate inference. To this end, we develop a framework to disrupt classical MVG inference in order to help decisionmakers understand the nature of indiscriminate adversarial disruption in this setting, explore the costs associated with said disruption, and spur inquiry about defensive measures.

More specifically, we consider a decisionmaker that utilizes the conditional distribution of a MVG for a variety of inference tasks that are required for subsequent decision-making.
However, an attacker intercepts the true data and corrupts it such that these inferences are thwarted and quality decision-making is impeded. We assume the attacker is motivated to maximally disrupt the decisionmaker's conditional distribution while minimizing their own risk of detection.
We quantify disruption using the Kullback-Leibler (KL) divergence between the conditional distributions generated under the corrupted and true data.
To quantify detection risk, we measure how probable the corrupted data is compared to its marginal mode.

Within this framework, where an attacker wishes to indiscriminately disrupt inference, balancing two separate objectives (disruption and detection), we construct solutions to the attacker's optimization problem in both a white-box (complete) and grey-box (incomplete) knowledge setting \citep{rios2023adversarial}. When the attacker has incomplete knowledge of the decisionmaker's model, he utilizes a Bayesian framework to quantify uncertainty about the unknown model parameters. Depending upon their structure, each of these problems require distinct solution methods. Such methods are explored herein and analysis is provided to determine their applicability.

Akin to the pioneering work of \citet{goodfellow2014explaining}, our focus is not strictly methodological. AML attacks are practically relevant insofar as they affect model behavior in application. Therefore, the impact of these attacks is evaluated within three applied examples characterized by distinct models, each expressible as an MVG. In particular, we explore the efficacy and flexibility of our approach in attacking an explicit MVG, a linear regression model with Gaussian error, and a linear Gaussian state-space model (LG-SSM). These models are respectively utilized in the distinct settings of real estate evaluation, interest rate estimation, and signals processing. For each of these applied examples, we explore the impacts of our attacks globally and on specified estimation tasks. In general, the attacks are shown to substantively affect the quality of the decisionmaker's inference; however, the fidelity of the attacker's knowledge is also shown to affect attack efficacy. These findings underscore the importance of operational security in applied probabilistic models. Since uncertainty complicates the search for an optimal attack, obfuscation of the model's parameterization may be a defense in its own right. Moreover, in some instances, corruption of subsets of the data is found to provide little benefit to the attacker; the impetus of this behavior in application is considered as well.

The remainder of this manuscript is structured as follows. Section \ref{sec2} formalizes the attacker's problems, whereas Sections \ref{sec:SolveWB} and \ref{sec:SolveGB} analyze their structural properties and explore potential solution methods. Section \ref{sec5} considers three applied examples to gain insight into attack efficacy and their associated effects on decisionmaker inference. Section \ref{sec6} provides concluding remarks and directions for future inquiry. 



\section{Disrupting Conditional Inference on MVGs} \label{sec2}


\subsection{Setup}

In what follows, we formally develop the competitive interaction between decisionmaker and attacker. The following notation is utilized in the sequel: vectors and matrices are denoted in bold text, scalars in standard font, bracketed subscripts represent a block of a vector/matrix, and unbracketed subscripts denote different variables or parameters of the same type. Moreover, when bracketed subscripts and matrix inversion (transposition) are used simultaneously, operations are performed on the block matrix itself.

\subsection{Decisionmaker Inference}

Assume a decisionmaker models $\left( {\mathbf{Y} \in \mathbb{R}^{|\mathbf{Y}|}, \mathbf{Z} \in \mathbb{R}^{|\mathbf{Z}|}}\right)\sim \mathcal{P}_{\mathbf{Y}\mathbf{Z}}=\mathcal{N}(\boldsymbol\mu, \boldsymbol\Sigma)$ with

$$ 
\boldsymbol\mu = \begin{bmatrix} \boldsymbol\mu_{[\mathbf{Y}]} \\ \boldsymbol\mu_{[\mathbf{Z}]} \end{bmatrix}, \quad 
\boldsymbol\Sigma = \begin{bmatrix} \boldsymbol\Sigma_{[\mathbf{Y}\mathbf{Y}]} & \boldsymbol\Sigma_{[\mathbf{Y}\mathbf{Z}]}\\
\boldsymbol\Sigma_{[\mathbf{Z}\mathbf{Y}]} & \boldsymbol\Sigma_{[\mathbf{Z}\mathbf{Z}]} \end{bmatrix}.
$$

\noindent Under the mild assumption that $\boldsymbol\Sigma$ is positive definite, the decisionmaker uses $\mathbf{Y}|\mathbf{z} \sim \mathcal{N}(\boldsymbol\mu_{\mathbf{Y}|\mathbf{z}}, \boldsymbol\Sigma_{\mathbf{Y}|\mathbf{z}})$ such that

\begin{subequations}
\begin{align}
    \boldsymbol\mu_{\mathbf{Y}|\mathbf{z}} &= \boldsymbol\mu_{[\mathbf{Y}]} + \boldsymbol\Sigma_{[\mathbf{Y} \mathbf{Z}]} \boldsymbol\Sigma^{-1}_{[\mathbf{Z} \mathbf{Z}]}(\mathbf{z}-\boldsymbol\mu_{[\mathbf{Z}]}), \\
    \boldsymbol\Sigma_{\mathbf{Y}|\mathbf{z}} & = \boldsymbol\Sigma_{[\mathbf{Y} \mathbf{Y}]} - \boldsymbol\Sigma_{[\mathbf{Y} \mathbf{Z}]} \boldsymbol\Sigma^{-1}_{[\mathbf{Z} \mathbf{Z}]} \boldsymbol\Sigma_{[\mathbf{Z} \mathbf{Y}]} , \label{eqcondK} 
\end{align}
\end{subequations}

\noindent to make inferences about $\mathbf{Y}$ or a subset of its variables. Note that $\boldsymbol\Sigma_{\mathbf{Y} | \mathbf{z}}$ is the Schur complement of $\boldsymbol\Sigma_{[\mathbf{Z}\mathbf{Z}]}$ in $\boldsymbol\Sigma$, i.e., $\boldsymbol\Sigma / \boldsymbol\Sigma_{[\mathbf{Z}\mathbf{Z}]}$. 

For notational convenience in subsequent derivations, we utilize the canonical form \citep[i.e., see page 609 of][]{koller2009probabilistic} wherein $ \mathcal{N}(\boldsymbol\mu, \boldsymbol\Sigma)=\mathcal{C}(\boldsymbol\Lambda,\boldsymbol\eta, \xi)$ such that

\begin{align*}
    \boldsymbol\Lambda & = \boldsymbol\Sigma^{-1},  \\
    \boldsymbol\eta &= \boldsymbol\Sigma^{-1} \boldsymbol\mu, \\
    \xi &= -\frac{1}{2}\boldsymbol\mu^T  \boldsymbol\Sigma^{-1} \boldsymbol\mu - \ln\left((2\pi)^{N/2} \det(\boldsymbol\Sigma)^{1/2}\right).
\end{align*}

\noindent The canonical-form representation of $\mathcal{P}_{\mathbf{Y}|\mathbf{z}}$ is $\mathcal{C}(\boldsymbol\Lambda_{\mathbf{Y}|\mathbf{z}},\boldsymbol\eta_{\mathbf{Y}|\mathbf{z}}, \xi_{\mathbf{Y}|\mathbf{z}})$ where

\begin{align*}
    \boldsymbol\Lambda_{\mathbf{Y}|\mathbf{z}} & = \boldsymbol\Lambda_{[\mathbf{Y}\mathbf{Y}]},  \\
    \boldsymbol\eta_{\mathbf{Y}|\mathbf{z}} &= \boldsymbol\eta_{\mathbf{[Y]}} - \boldsymbol\Lambda_{[\mathbf{Y}\mathbf{Z}]} \mathbf{z},  \\
    \xi_{\mathbf{Y}|\mathbf{z}} &= \xi -\boldsymbol\eta_{\mathbf{[Z]}}^T \mathbf{z} - \frac{1}{2}\mathbf{z}^T \boldsymbol\Lambda_{[\mathbf{Z}\mathbf{Z}]} \mathbf{z}.
\end{align*}

\noindent and, by properties of the Schur complement, we have $\boldsymbol\Lambda_{\mathbf{Y}|\mathbf{z}} = \boldsymbol\Lambda_{[\mathbf{Y}\mathbf{Y}]} = \boldsymbol\Sigma^{-1}_{\mathbf{Y}|\mathbf{z}}$, implying, of course, that $\boldsymbol\Lambda^{-1}_{\mathbf{Y}|\mathbf{z}} = \boldsymbol\Lambda^{-1}_{[\mathbf{Y}\mathbf{Y}]} = \boldsymbol\Sigma_{\mathbf{Y}|\mathbf{z}}$.

\subsection{Corrupting the Conditional Distribution}

The attacker aims to corrupt the observed data such that the conditional distribution over the unobserved data is maximally disrupted while ensuring their assault is undetected. We assume that the attacker is able to intercept the true data, $\mathbf{z}'$, and modify it to $\mathbf{z}$ to achieve their aims. Formally, the attacker wishes to find $\mathbf{z} \in \mathcal{Z}$ that causes maximal disruption with minimal risk wherein $\mathcal{Z}$ is a bounded feasible attack region with non-empty interior that accounts for decisionmaker security and attack limitations. We consider two variants of this distribution-disruption problem that alternatively assume the attacker has complete or partial knowledge about $\mathcal{P}_{\mathbf{Y}\mathbf{Z}}$ (i.e., white- and grey-box settings). Herein, we formulate each of these settings as multiobjective optimization problems. 




The attacker measures the corruption of $\mathcal{P}_{\mathbf{Y}|\mathbf{z}}$ from $\mathcal{P}_{\mathbf{Y}|\mathbf{z}'}$ using the Kullback-Leibler divergence. Since both distributions are MVGs,



\begin{align*}
D_{KL}(\mathcal{P}_{\mathbf{Y}|\mathbf{z}'}||\mathcal{P}_{\mathbf{Y}|\mathbf{z}}) &= \frac{1}{2} \left(\boldsymbol\mu_{\mathbf{Y}|\mathbf{z}'} - \boldsymbol\mu_{\mathbf{Y}|\mathbf{z}} \right)^T \boldsymbol\Sigma_{\mathbf{Y}|\mathbf{z}}^{-1} \left(\boldsymbol\mu_{\mathbf{Y}|\mathbf{z}'} - \boldsymbol\mu_{\mathbf{Y}|\mathbf{z}}\right)\\ & \qquad + \frac{1}{2}\left(Tr\left(\boldsymbol\Sigma^{-1}_{\mathbf{Y}|\mathbf{z}} \boldsymbol\Sigma_{\mathbf{Y}|\mathbf{z}'}\right) + \ln\left( \frac{|\boldsymbol\Sigma_{\mathbf{Y}|\mathbf{z}}|}{|\boldsymbol\Sigma_{\mathbf{Y}|\mathbf{z}'}|}  \right) - |\mathbf{Y}| \right).
\end{align*}

\noindent However, for every attack $\mathbf{z}$, $\boldsymbol\Sigma_{\mathbf{Y}|\mathbf{z}} = \boldsymbol\Sigma_{\mathbf{Y}|\mathbf{z}'}$ by Equation \eqref{eqcondK}. Therefore, $\frac{|\boldsymbol\Sigma_{\mathbf{Y}|\mathbf{z}}|}{|\boldsymbol\Sigma_{\mathbf{Y}|\mathbf{z}'}|}=1$ and $Tr(\boldsymbol\Sigma^{-1}_{\mathbf{Y}|\mathbf{z}} \boldsymbol\Sigma_{\mathbf{Y}|\mathbf{z}'})={|\mathbf{Y}|}$, further implying that 


\begin{align*}
D_{KL}(\mathcal{P}_{{\mathbf{Y}|\mathbf{z}'}}||\mathcal{P}_{\mathbf{Y}|\mathbf{z}}) &= \frac{1}{2} \left(\boldsymbol\mu_{\mathbf{Y}|\mathbf{z}'} - \boldsymbol\mu_{\mathbf{Y}|\mathbf{z}}\right)^T \boldsymbol\Sigma_{\mathbf{Y}|\mathbf{z}'}^{-1} \left(\boldsymbol\mu_{\mathbf{Y}|\mathbf{z}'} - \boldsymbol\mu_{\mathbf{Y}|\mathbf{z}}\right) .
\end{align*}

\noindent Note that the disruption measurement is symmetric, $D_{KL}(\mathcal{P}_{{\mathbf{Y}|\mathbf{z}'}}||\mathcal{P}_{\mathbf{Y}|\mathbf{z}}) = D_{KL}(\mathcal{P}_{{\mathbf{Y}|\mathbf{z}}}||\mathcal{P}_{\mathbf{Y}|\mathbf{z}'})$ and has decomposed to a Mahalanobis distance. 
Furthermore, since $\mathbf{z}$ does not affect $\boldsymbol\mu_{\mathbf{Y}|\mathbf{z}'}$,  Property \ref{prop:KLinz} derives $D_{KL}(\mathcal{P}_{{\mathbf{Y}|\mathbf{z}'}}||\mathcal{P}_{\mathbf{Y}|\mathbf{z}}) $ as a function of $\mathbf{z}$. 

\begin{property} \small \label{prop:KLinz} If $\mathbf{z}'$ is the true evidence and $\mathbf{z}$ the corrupted evidence, the KL divergence between the decisionmaker's true and corrupted conditional distributions is 
$$D_{KL}(\mathcal{P}_{{\mathbf{Y}|\mathbf{z}'}}||\mathcal{P}_{\mathbf{Y}|\mathbf{z}}) = \frac{1}{2}\left( \mathbf{z}^T \mathbf{Q} \mathbf{z} + \mathbf{v}^T \mathbf{z} + c \right)$$

\noindent  where

\begin{align*}
\mathbf{Q} &= \boldsymbol\Lambda_{[\mathbf{Y}\mathbf{Z}]}^T \boldsymbol\Lambda^{-1}_{{[\mathbf{Y}\mathbf{Y}]}} \boldsymbol\Lambda_{[\mathbf{Y}\mathbf{Z}]}, \\
\mathbf{v} &= 2\left( \boldsymbol\Lambda_{[\mathbf{Y}\mathbf{Z}]}^T \boldsymbol\mu_{\mathbf{Y}|\mathbf{z}'} - \boldsymbol\Lambda^T_{[\mathbf{Y}\mathbf{Z}]} \boldsymbol\Lambda^{-1}_{{[\mathbf{Y}\mathbf{Y}]}} \boldsymbol\eta_{[\mathbf{Y}]} \right), \\
c &= \boldsymbol\mu_{\mathbf{Y}|\mathbf{z}'}^T \boldsymbol\Lambda_{{[\mathbf{Y}\mathbf{Y}]}} \boldsymbol\mu_{\mathbf{Y}|\mathbf{z}'} - 2 \boldsymbol\eta^T_{[\mathbf{Y}]}\boldsymbol\mu_{\mathbf{Y}|\mathbf{z}'} + \boldsymbol\eta^T_{[\mathbf{Y}]} \boldsymbol\Lambda^{-1}_{{[\mathbf{Y}\mathbf{Y}]}} \boldsymbol\eta_{[\mathbf{Y}]}.
\end{align*}
\end{property}

\begin{proof} \small

The Kullback-Leibler divergence expands to 
\begin{align*}
    D_{KL}(\mathcal{P}_{{\mathbf{Y}|\mathbf{z}'}}||\mathcal{P}_{\mathbf{Y}|\mathbf{z}}) &= \frac{1}{2}\left(\boldsymbol\mu_{\mathbf{Y}|\mathbf{z}'}^T \boldsymbol\Sigma_{\mathbf{Y}|\mathbf{z}}^{-1}\boldsymbol\mu_{\mathbf{Y}|\mathbf{z}'} - 2\boldsymbol\mu_{\mathbf{Y}|\mathbf{z}}^T \boldsymbol\Sigma_{\mathbf{Y}|\mathbf{z}}^{-1}\boldsymbol\mu_{\mathbf{Y}|\mathbf{z}'} + \boldsymbol\mu_{\mathbf{Y}|\mathbf{z}}^T \boldsymbol\Sigma_{\mathbf{Y}|\mathbf{z}}^{-1}\boldsymbol\mu_{\mathbf{Y}|\mathbf{z}} \right)  
\end{align*}

\noindent and the relationships between the canonical- and moment-form MVG representations imply
 


\begin{align*}
  - 2\boldsymbol\mu_{\mathbf{Y}|\mathbf{z}}^T \boldsymbol\Sigma_{\mathbf{Y}|\mathbf{z}}^{-1}\boldsymbol\mu_{\mathbf{Y}|\mathbf{z}'} & = -2 \boldsymbol\eta^T_{[\mathbf{Y}]}\boldsymbol\mu_{\mathbf{Y}|\mathbf{z}'} + 2 \mathbf{z}^T \boldsymbol\Lambda^T_{[\mathbf{Y}\mathbf{Z}]}\boldsymbol\mu_{\mathbf{Y}|\mathbf{z}'},
\end{align*}

\noindent and 

\begin{align*}
    \boldsymbol\mu_{\mathbf{Y}|\mathbf{z}}^T \boldsymbol\Sigma_{\mathbf{Y}|\mathbf{z}}^{-1}\boldsymbol\mu_{\mathbf{Y}|\mathbf{z}} = \boldsymbol\eta^T_{[\mathbf{Y}]}\boldsymbol\Sigma_{\mathbf{Y}|\mathbf{z}}\boldsymbol\eta_{[\mathbf{Y}]} -2\mathbf{z}^T \boldsymbol\Lambda_{[\mathbf{Y}\mathbf{Z}]}^T \boldsymbol\Sigma_{\mathbf{Y}|\mathbf{z}}\boldsymbol\eta_{[\mathbf{Y}]} + \mathbf{z}^T \boldsymbol\Lambda^T_{[\mathbf{Y}\mathbf{Z}]}\boldsymbol\Sigma_{\mathbf{Y}|z} \boldsymbol\Lambda_{[\mathbf{Y}\mathbf{Z}]} \mathbf{z}.
\end{align*}

\noindent Collecting terms and substituting $\boldsymbol\Sigma_{\mathbf{Y}|\mathbf{z}} = \boldsymbol\Lambda^{-1}_{[\mathbf{Y}\mathbf{Y}]} $ yields $\mathbf{Q}$, $\mathbf{v}$ and $c$, as previously defined. 

\end{proof} \normalsize


\noindent Based on these results, maximizing disruption $D_{KL}(\mathcal{P}_{{\mathbf{Y}|\mathbf{z}'}}||\mathcal{P}_{\mathbf{Y}|\mathbf{z}})$ is equivalent to maximizing 

$$
\phi_1(\mathbf{z}) = \mathbf{z}^T \mathbf{Q} \mathbf{z} + \mathbf{v}^T \mathbf{z}. 
$$



Additionally, the attacker assumes less risk if $\mathbf{z}$ is deemed to be a sufficiently probable observation based upon $\mathcal{P}_{\mathbf{Y}\mathbf{Z}}$. We measure detection risk by the log-ratio of the marginal density of $\mathbf{Z}$ evaluated at $\mathbf{z}$ and the mode of $\mathbf{Z}$.
By properties of the MVG, we have $\mathbf{Z} \sim \mathcal{N}(\boldsymbol\mu_{[\mathbf{Z}]}, \boldsymbol\Sigma_{[\mathbf{Z}\mathbf{Z}]})$, and the log-ratio of marginal densities between $\mathbf{z}$ and the marginal mode of $\mathbf{Z}$ is

\begin{align*}
\ln \left( \frac{f_{\mathbf{Z}}(\mathbf{z})}{f_{\mathbf{Z}}(\boldsymbol\mu_{[\mathbf{Z}]})} \right) &= \ln \left(\frac{\exp\left(-\frac{1}{2}\left(\mathbf{z}-\boldsymbol\mu_{[\mathbf{Z}]}\right)^T \boldsymbol\Sigma^{-1}_{[\mathbf{Z}\mathbf{Z}]} \left(\mathbf{z}-\boldsymbol\mu_{[\mathbf{Z}]}\right)\right)}{\exp\left(-\frac{1}{2}\left(\boldsymbol\mu_{[\mathbf{Z}]}-\boldsymbol\mu_{[\mathbf{Z}]}\right)^T \boldsymbol\Sigma^{-1}_{[\mathbf{Z}\mathbf{Z}]} \left(\boldsymbol\mu_{[\mathbf{Z}]}-\boldsymbol\mu_{[\mathbf{Z}]}\right)\right)} \right),\\
&= -\frac{1}{2}\left(\mathbf{z}-\boldsymbol\mu_{[\mathbf{Z}]}\right)^T \boldsymbol\Sigma^{-1}_{[\mathbf{Z}\mathbf{Z}]} \left(\mathbf{z}-\boldsymbol\mu_{[\mathbf{Z}]}\right), \\
&= -\frac{1}{2}\left( \mathbf{z}^T \boldsymbol\Sigma^{-1}_{[\mathbf{Z}\mathbf{Z}]}\mathbf{z} - 2\mathbf{z}^T \boldsymbol\Sigma^{-1}_{[\mathbf{Z}\mathbf{Z}]}\boldsymbol\mu_{[\mathbf{Z}]} +  \boldsymbol\mu_{[\mathbf{Z}]}^T \boldsymbol\Sigma^{-1}_{[\mathbf{Z}\mathbf{Z}]}\boldsymbol\mu_{[\mathbf{Z}]}\right).
\end{align*}


\noindent Accounting for constants, we see that optimizing the log-ratio of marginal densities is equivalent to optimizing

$$
\phi_2(\mathbf{z})=- \mathbf{z}^T \boldsymbol\Sigma^{-1}_{[\mathbf{Z}\mathbf{Z}]}\mathbf{z}+ 2\mathbf{z}^T  \boldsymbol\Sigma^{-1}_{[\mathbf{Z}\mathbf{Z}]}\boldsymbol\mu_{[\mathbf{Z}]} . 
$$

\subsection{Attacker Problem Formulations} \label{secProbForm}


In the white-box (WB) setting, the attacker has  complete knowledge about $\mathcal{P}_{\mathbf{Y}\mathbf{Z}}$, and finds their optimal attack by optimizing $w_1 \phi_1(\mathbf{z}) +  w_2 \phi_2(\mathbf{z})$ as expressed below in Problem \textbf{WB},  


\begin{align*}
\textbf{WB}: \ & \max_{\mathbf{z} \in \mathcal{Z}} \  \mathbf{z}^T\left(w_1\mathbf{Q}- w_2 \boldsymbol\Sigma^{-1}_{[\mathbf{Z}\mathbf{Z}]}\right) \mathbf{z} + \mathbf{z}^T\left(w_1 \mathbf{v} + 2w_2\boldsymbol\Sigma^{-1}_{[\mathbf{Z}\mathbf{Z}]}\boldsymbol\mu_{[\mathbf{Z}]}\right), 
\end{align*}

\noindent where $w_1 =\frac{u_1}{|\phi^*_1|}$ and $w_2 = \frac{u_2}{|\phi_2^*|}$ are normalized weights,  $u_k \in (0,1)$ are raw weights summing to one and $\phi^*_k$ are single-objective optimal values over $\mathcal{Z}$ (i.e., with $w_1$ and $w_2$ alternatively set to zero and one). Such normalization enables each component objective to be measured on a comparable scale, thereby ensuring numerical stability and providing a more intuitive meaning of Problem WB's objective function. This formulation is explored in greater detail in Section \ref{sec:SolveWB}.


Alternatively, if the attacker has partial knowledge of $\mathcal{P}_{\mathbf{Y}\mathbf{Z}}$, their beliefs may be specified in a Bayesian manner such that their uncertainty about the unknown parameters is quantified by an associated prior distribution. Under such conditions, Problem WB can be readily recast to the grey-box setting as



\begin{align*}
\textbf{GB}: \ \max_{\mathbf{z} \in \mathcal{Z}} \  \mathbb{E} \left[  \mathbf{z}^T\left(w_1\mathbf{Q}- w_2 \boldsymbol\Sigma^{-1}_{[\mathbf{Z}\mathbf{Z}]}\right) \mathbf{z} + \mathbf{z}^T\left(w_1 \mathbf{v} + 2w_2\boldsymbol\Sigma^{-1}_{[\mathbf{Z}\mathbf{Z}]}\boldsymbol\mu_{[\mathbf{Z}]}\right) \right]
\end{align*}

\noindent where the expectation is taken with respect to the prior distribution over the unknown parameters. However, as will be discussed in Section \ref{sec:SolveGB}, this notationally simple modification belies a problem of substantially greater difficulty. 

Notably, formulating the attacker's objective using the KL divergence and log-ratio of marginal densities is but one of myriad alternatives. Alternative statistical distances could be used to assess the difference between $\mathcal{P}_{{\mathbf{Y}|\mathbf{z}'}}$ and $\mathcal{P}_{{\mathbf{Y}|\mathbf{z}}}$, and distinct penalties could be assessed to enforce plausibility. However, as will be shown in subsequent sections, our formulation strikes a tailored balance between analytic expressiveness and computational difficulty.

\section{Solving the White-box Problem} \label{sec:SolveWB}

Within this section, we explore the structure of the white-box problem and then discuss how it may be solved. 


\subsection{Problem Structure Insights}


To build insight, it is necessary to first recognize the following elementary properties. 

\begin{property} \small \label{propQ_PD}
$\mathbf{Q}$ is positive semi-definite.
\end{property}
\begin{proof} \small
Since $\boldsymbol{\Lambda}$ is an invertible precision matrix, $\boldsymbol{\Lambda}$ is positive definite. $\boldsymbol{\Lambda}_{[\mathbf{Y}\mathbf{Y}]}$ is a principal submatrix of $\boldsymbol{\Lambda}$ and is also positive definite. Positive definiteness is preserved under inversion, implying $\boldsymbol{\Lambda}^{-1}_{[\mathbf{Y}\mathbf{Y}]}$ is positive definite. Thus, $\boldsymbol{\Lambda}^{-1}_{[\mathbf{Y}\mathbf{Y}]}$ is also positive semi-definite and, since 

$$\mathbf{z}^T\mathbf{Q}\mathbf{z} = \mathbf{z}^T\left(\boldsymbol\Lambda_{[\mathbf{Y}\mathbf{Z}]}^T \boldsymbol\Lambda^{-1}_{{[\mathbf{Y}\mathbf{Y}]}}\boldsymbol\Lambda_{[\mathbf{Y}\mathbf{Z}]}\right)\mathbf{z} =\left(\boldsymbol\Lambda_{[\mathbf{Y}\mathbf{Z}]} \mathbf{z}\right)^T \boldsymbol\Lambda^{-1}_{{[\mathbf{Y}\mathbf{Y}]}}\left(\boldsymbol\Lambda_{[\mathbf{Y}\mathbf{Z}]} \mathbf{z}\right) \ge 0, \ \forall z \in \mathbb{R}^{|\mathbf{Z}|} ,$$

\noindent $\mathbf{Q}$ is positive semi-definite.  
\end{proof} \normalsize

\begin{property} \small \label{propSigma_PD}
$\boldsymbol{\Sigma}^{-1}_{[\mathbf{Z}\mathbf{Z}]}$ is positive definite. 
\end{property}
\begin{proof} \small 
$\boldsymbol{\Sigma}_{[\mathbf{Z}\mathbf{Z}]}$ is a principal submatrix of $\boldsymbol{\Sigma}$ and thus positive definite. Since positive definiteness is preserved under inversion, $\boldsymbol{\Sigma}^{-1}_{[\mathbf{Z}\mathbf{Z}]}$ is positive definite.
\end{proof} \normalsize

From this characterization, it is apparent that $\phi_1$ is convex and $\phi_2$ is concave. Since Problem WB's objective function equals $w_1\phi_1(\mathbf{z}) + w_2\phi_2(\mathbf{z})$, it may thus be concave, convex or neither convex nor concave. Letting $\{\rho_m\}_{m=1}^{|\mathbf{Z}|}$, $\{\zeta_m\}_{m=1}^{|\mathbf{Z}|}$, and $\{\lambda_m\}_{m=1}^{|\mathbf{Z}|}$ be the respective eigenvalues of $\mathbf{Q}$, $\boldsymbol\Sigma_{[\mathbf{Z}\mathbf{Z}]}$, and $w_1\mathbf{Q}- w_2 \boldsymbol\Sigma^{-1}_{[\mathbf{Z}\mathbf{Z}]}$ arranged in non-ascending order, the objective function's convexity can be characterized as follows.  

\begin{property} \small \label{propWB_concave}
Problem WB's objective function is concave if for every $i \in \{1, \cdots, |\mathbf{Z}|\}$, there exists $m,n \ge 1$ such that $i=m+n-1$, $1 \le m+n-1 \le |\mathbf{Z}|$, and $w_1\rho_m - w_2 \zeta_{n} \le 0$.
\end{property}
\begin{proof} \small
By Properties \ref{propQ_PD} and \ref{propSigma_PD}, $\rho_m\ge0$, and $\zeta_m >0, \ \forall m$. The non-negativity of the $w_k$-values implies the eigenvalues of $w_1\mathbf{Q}$ and $-w_2\boldsymbol\Sigma^{-1}_{[\mathbf{Z}\mathbf{Z}]}$ are $\{w_1\rho_m\}_{m=1}^{|\mathbf{Z}|}$ and $\{-w_2\zeta_m\}_{m=1}^{|\mathbf{Z}|}$ such that the former are non-negative and the latter are strictly negative. The symmetry of $\mathbf{Q}$ and $\boldsymbol\Sigma^{-1}_{[\mathbf{Z}\mathbf{Z}]}$ implies $w_1\mathbf{Q}$ and $-w_2\boldsymbol\Sigma^{-1}_{[\mathbf{Z}\mathbf{Z}]}$ are Hermitian and, by Weyl's inequalities, we have
$$
\lambda_{m+n-1} \le w_1\rho_m - w_2 \zeta_{n}
$$

\noindent whenever $1 \le m+n-1 \le |\mathbf{Z}|$. Thus, if for every $i \in \{1, \cdots, |\mathbf{Z}|\}$ there exists $m,n\ge 1$ such that $i=m+n-1$ with $1\le m+n-1 \le |\mathbf{Z}|$ and $w_1\rho_m - w_2 \zeta_{n} \le 0$, all $\lambda_{i}$ are non-positive, $w_1\mathbf{Q}- w_2 \boldsymbol\Sigma^{-1}_{[\mathbf{Z}\mathbf{Z}]}$ is negative semidefinite, and $w_1\phi_1(\mathbf{z}) + w_2\phi_2(\mathbf{z})$ is a concave function.
\end{proof} \normalsize

\begin{property} \small \label{propWB_convex}
Problem WB's objective function is convex if for every $i \in \{1, \cdots, |\mathbf{Z}|\}$, there exists $m,n \ge 1$ such that $i=m+n-|\mathbf{Z}|$, $1\le m+n - |\mathbf{Z}| \le |\mathbf{Z}|$, and $0 \le w_1\rho_m - w_2 \zeta_{n}$.
\end{property}
\begin{proof} \small
Since $w_1\mathbf{Q}$ and $-w_2\boldsymbol\Sigma^{-1}_{[\mathbf{Z}\mathbf{Z}]}$ are Hermitian matrices, Weyl's inequalities imply that 

$$
w_1\rho_m - w_2 \zeta_{n} \le \lambda_{m+n-|\mathbf{Z}|}
$$

\noindent whenever $1\le m+n - |\mathbf{Z}| \le |\mathbf{Z}|$. Therefore, if for every $i \in \{1, \cdots, |\mathbf{Z}|\}$ there exists $m, n \ge 1$ such that $i=m+n-|\mathbf{Z}|$ with $1\le m+n - |\mathbf{Z}| \le |\mathbf{Z}|$ and $0 \le w_1\rho_m - w_2 \zeta_{n}$, all $\lambda_{i}$ are non-negative,  $w_1\mathbf{Q}- w_2 \boldsymbol\Sigma^{-1}_{[\mathbf{Z}\mathbf{Z}]}$ is positive semidefinite and $w_1\phi_1(\mathbf{z}) + w_2\phi_2(\mathbf{z})$ is a convex function.
\end{proof} \normalsize


Moreover, recalling that $w_k = \frac{u_k}{|\phi^*_k|}$, Properties \ref{propWB_concave} and \ref{propWB_convex} can be directly extended to $u_1$ since $u_2=1-u_1$. That is, if for some $m$ and $n$, we have  $\frac{u_1}{|\phi^*_1|}\rho_m$ - $\frac{(1-u_1)}{|\phi^*_2|}\zeta_n \le 0$, Weyl's inequality can be rearranged such that

$$
u_1 \le \frac{\zeta_n}{|\phi^*_2|} \left(\frac{\rho_m}{|\phi^*_1|} + \frac{\zeta_n}{|\phi^*_2|}\right)^{-1}
$$.

\noindent Therefore, since the eigenvalues of $\mathbf{Q}$ and $\boldsymbol\Sigma^{-1}_{[\mathbf{Z}\mathbf{Z}]}$ are readily discernible, if the following bound holds, $w_1\phi_1(\mathbf{z})+w_2\phi_2(\mathbf{z})$ is a concave function:

\begin{align} \label{eqConcaveBound}
u_1\le u_1^{-}= \min  \left\{ u_1 \le \frac{\zeta_n}{|\phi^*_2|} \left(\frac{\rho_m}{|\phi^*_1|} + \frac{\zeta_n}{|\phi^*_2|}\right)^{-1}: \ 1\le m+n-1 \le |\mathbf{Z}| \right\}.
\end{align} 

\noindent A similar sufficient condition can be derived for the function being convex:

\begin{align} \label{eqConvexBound}
u_1 \ge u_1^{+} = \max  \left\{ u_1 \le \frac{\zeta_n}{|\phi^*_2|} \left(\frac{\rho_m}{|\phi^*_1|} + \frac{\zeta_n}{|\phi^*_2|}\right)^{-1}: 1 \le \ m+n-|\mathbf{Z}| \le |\mathbf{Z}| \right\}.
\end{align}

\begin{figure}
    \centering
    \begin{subfigure}[b]{0.4\textwidth}
         \centering
         \includegraphics[width=\textwidth]{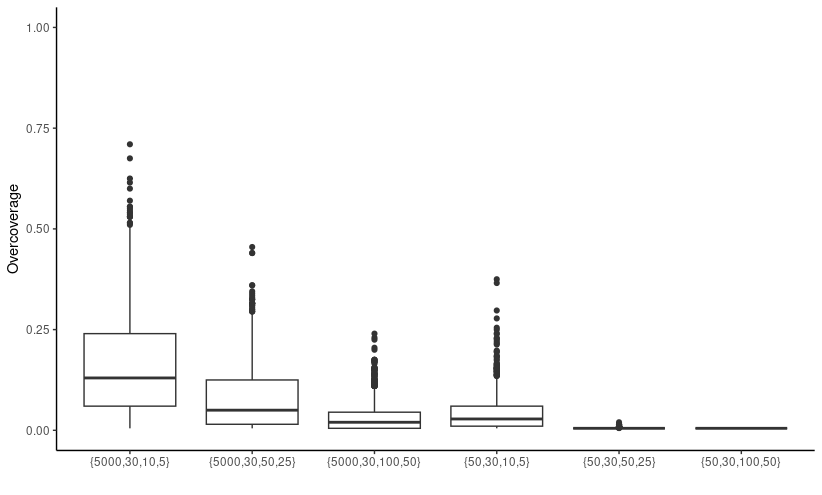}
         \caption{Normal-Transpose-Multiplication}
         \label{fig:normOvercoverage}
     \end{subfigure}
     \begin{subfigure}[b]{0.4\textwidth}
         \centering
         \includegraphics[width=\textwidth]{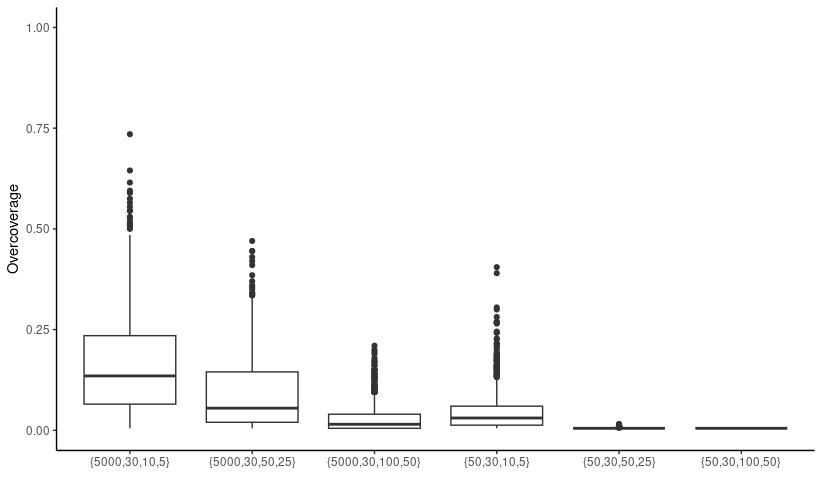}
         \caption{Inverse-Wishart}
         \label{fig:IWOvercoverage}
     \end{subfigure}
    \caption{\centering Overcoverage of the $u_1$ interval inducing a neither-convex-nor-concave objective function by covariance matrix sampling and $\{|\mathbf{Y}| +|\mathbf{Z}|, |\mathbf{Z}|, |\phi^*_1|, |\phi^*_2| \}$}
    \label{fig:boundOvercoverage}
\end{figure}

However, if $u_1^- < u_1 < u_1^+$, determining convexity of the objective function requires further analysis. In other words, the bounds are conservative. We assess this conservativeness empirically in Figure \ref{fig:boundOvercoverage}. We estimate the true convexity-transition points through brute force searches within 0.005 of its true value for multiple values of $|\mathbf{Y}| +|\mathbf{Z}|$, $|\mathbf{Z}|$, $|\phi^*_1|$,  and $|\phi^*_2|$; these brute force approximations are respectively denoted as $\tilde{u}_1^-$ and $\tilde{u}_1^+$. For each setting, we form 1000  $\boldsymbol\Sigma$ by (1) generating a matrix of random entries from a standard normal, multiplying this by its transpose, and checking for positive definiteness; and (2) sampling from $\mathcal{IW}(\mathbf{I}, |\mathbf{Z}|+|\mathbf{Y}|)$ where $\mathbf{I}$ is the identity matrix having appropriate dimension. Figure \ref{fig:boundOvercoverage} shows the overcoverage of our bounds against the interval for which the objective function is truly neither convex nor concave, i.e., $(u_1^+ - u_1^-) - (\tilde{u}_1^+ - \tilde{u}_1^-)$.

A few notable patterns emerge from inspection of Figure \ref{fig:boundOvercoverage}. First, the bounds' overcoverage tends to decrease with the size of the covariance matrix. Moreover, if $|\phi^*_1|$ or $|\phi^*_2|$ increase in isolation, the degree of overcoverage increases, and the overcoverage is smaller when the values are similarly sized. However, we note that as  $|\phi^*_1|$ and $|\phi^*_2|$ increase in isolation, the tightness of $u_1^+$ and $u_1^-$ to the convexity-transition point increases; for example, a greater $|\phi^*_1|$ tended to push $u_1^+$ closer to the true transition. Moreover, in all cases examined, the median overcoverage is less than 0.15. 

\subsection{Solution Techniques for the White-box Problem} \label{sec:SolvingWB}

The convexity of $\mathcal{Z}$ and $w_1\phi_1(\mathbf{z})+w_2\phi_2(\mathbf{z})$ determines how the attacker should solve Problem WB. 
If $\mathcal{Z}$ is a polyhedron formed by affine constraints and Property \ref{propWB_concave} is satisfied, the attacker aims to solve a convex quadratic program which can be approached, e.g., as a linear complementarity problem \citep{bazaraa2013nonlinear}. Alternatively, if $\mathcal{Z}$ is not a polyhedron but remains convex, interior-point methods may be leveraged instead \citep{boyd2004convex}. Optimal attacks are easy to identify.

Conversely, optimal attacks are more challenging to locate in other settings. Even with convex $\mathcal{Z}$, if Property \ref{propWB_convex} is satisfied, the instance is generally NP-Hard; the same applies if $w_1\phi_1(\mathbf{z})+w_2\phi_2(\mathbf{z})$ is neither convex nor concave. In the former setting, it is known that the optimal solution lies on an extreme point of $\mathcal{Z}$, a fact that is leveraged in the concave programming methods detailed by \citet{pardalos1986methods}. However, in the latter setting, \citet{pardalos1991global} explain that optimal solutions needn't lie on an extreme point, e.g., they may lie on a face of a polyhedron. Under such conditions, specialized algorithms exist via reformalization-linearization \citep[e.g., see][]{sherali1995reformulation} and semi-definite programming relaxations \citep[e.g., see][]{burer2008finite}. For instances characterized by a neither-convex-nor-concave objective function and a non-convex $\mathcal{Z}$, the attacker must resort to heuristic methods (e.g., multi-start gradient searches) or global solvers \citep[e.g., see][]{sahinidis1996baron, bao2015global}. Many of these methods are available, via one implementation or another, within commercial and open-source optimization software. 

\section{Solving the Grey-box Problem} \label{sec:SolveGB}

Herein, we analyze the grey-box problem to determine how it may be solved. Thereafter, we discuss potential solution methods. 


\subsection{Problem Structure Insights}
{\color{black}
By the linearity of expectation, if the parameters' expected values can be identified, the properties derived for Problem WB extend naturally to Problem GB. Unfortunately, under a generic prior, one must rely upon simulation to approximate these expectations, implying convexity bounds on the objective function cannot always be identified analytically. However, for some priors, Inequalities \eqref{eqConcaveBound} and \eqref{eqConvexBound} can be readily extended to Problem GB.





Consider the normal-inverse-Wishart distribution as the attacker's prior where 

$$
(\boldsymbol\mu, \boldsymbol\Sigma) \sim \mathcal{NIW}(\boldsymbol\mu_0, \kappa, \boldsymbol\Psi\ , \nu)
$$

\noindent such that

\begin{align*}
\boldsymbol\mu &\sim \mathcal{N}\left(\boldsymbol\mu_0, \frac{1}{\kappa}\boldsymbol\Sigma \right), \\
\boldsymbol\Sigma &\sim \mathcal{IW}\left(\boldsymbol\Psi, \nu \right),
\end{align*}

\noindent $\mathcal{IW}$ is the inverse Wishart distribution over positive definite matrices in $\mathbb{R}^{|\mathbf{Y}|+|\mathbf{Z}|} \times \mathbb{R}^{|\mathbf{Y}|+|\mathbf{Z}|}$, and $\{\boldsymbol\mu_0, \kappa, \boldsymbol\Psi\ , \nu\}$ are the distribution's corresponding hyperparameters. Under these conditions, $\mathbb{E}\left[\mathbf{Q}\right]$ and $\mathbb{E}\left[\boldsymbol\Sigma^{-1}_{[\mathbf{Z}\mathbf{Z}]}\right]$ have closed form expressions.


\begin{property} \small \label{prop:EVGB}
If the attacker's prior is $(\boldsymbol\mu, \boldsymbol\Sigma) \sim \mathcal{NIW}(\boldsymbol\mu_0, \kappa, \boldsymbol\Psi\ , \nu) $ then
\begin{align*} 
\mathbb{E}\left[\mathbf{Q}\right]&=\left(\nu - |\mathbf{Y}|\right) \left(\boldsymbol\Omega_{[\mathbf{Z}\mathbf{Y}]}\boldsymbol\Omega^{-1}_{[\mathbf{Y}\mathbf{Y}]}\boldsymbol\Omega_{[\mathbf{Y}\mathbf{Z}]} \right) + |\mathbf{Y}| \Omega_{[\mathbf{Z}\mathbf{Z}]}  , \text{and}\\ \mathbb{E}\left[\boldsymbol\Sigma^{-1}_{[\mathbf{Z}\mathbf{Z}]}\right]&= \left(\nu - |\mathbf{Y}|\right)\boldsymbol\Psi^{-1}_{[\mathbf{Z}\mathbf{Z}]}
\end{align*} \color{black}
\noindent such that $\boldsymbol\Omega = \Psi^{-1}$.
\end{property}

\begin{proof}
Since $\boldsymbol\Sigma \sim \mathcal{IW}\left(\boldsymbol\Psi, \nu \right)$, we have $\boldsymbol\Lambda = \boldsymbol\Sigma^{-1} \sim \mathcal{W}(\boldsymbol\Omega, \nu)$ where $\mathcal{W}$ is the Wishart distribution. By \citet{ouellette1981schur}, Schur complements of Wishart matrices also have a Wishart distribution, implying 

$$\boldsymbol\Lambda/\boldsymbol\Lambda_{[\mathbf{Y}\mathbf{Y}]} = \boldsymbol\Lambda_{[\mathbf{Z}\mathbf{Z}]}- \boldsymbol\Lambda^T_{[\mathbf{Y}\mathbf{Z}]} \boldsymbol\Lambda^{-1}_{[\mathbf{Y}\mathbf{Y}]}\boldsymbol\Lambda_{[\mathbf{Y}\mathbf{Y}]} \sim \mathcal{W}(\boldsymbol\Omega/\boldsymbol\Omega_{[\mathbf{Y}\mathbf{Y}]}, \nu - |\mathbf{Y}|)$$

\noindent Likewise, since $\boldsymbol\Lambda$ is Wishart, $\boldsymbol\Lambda_{[\mathbf{Z}\mathbf{Z}]} \sim \mathcal{W}(\boldsymbol\Omega_{[\mathcal{Z}, \mathcal{Z}]}, \nu )$. Noting that $\mathbb{E}\left[\boldsymbol\Lambda/\boldsymbol\Lambda_{[\mathbf{Y}\mathbf{Y}]}\right] = \mathbb{E}\left[\boldsymbol\Lambda_{[\mathbf{Z}\mathbf{Z}]}\right] - \mathbb{E}[\mathbf{Q}]$, we have

$$
\mathbb{E}[\mathbf{Q}] = \left(\nu - |\mathbf{Y}|\right) \left(\boldsymbol\Omega_{[\mathbf{Z}\mathbf{Y}]}\boldsymbol\Omega^{-1}_{[\mathbf{Y}\mathbf{Y}]}\boldsymbol\Omega_{[\mathbf{Y}\mathbf{Z}]} \right) + |\mathbf{Y}| \Omega_{[\mathbf{Z}\mathbf{Z}]} 
$$
\color{black}




\noindent Finally, since $\boldsymbol\Sigma \sim \mathcal{IW}\left(\boldsymbol\Psi, \nu \right)$,  we necessarily have $\boldsymbol\Sigma_{[\mathbf{Z}\mathbf{Z}]} \sim \mathcal{IW}\left(\boldsymbol\Psi_{[\mathbf{Z}\mathbf{Z}]}, \nu - |\mathbf{Y}| \right)$, and $\boldsymbol\Sigma^{-1}_{[\mathbf{Z}\mathbf{Z}]} \sim \mathcal{W}\left(\boldsymbol\Psi^{-1}_{[\mathbf{Z}\mathbf{Z}]}, \nu - |\mathbf{Y}| \right)$. Thus, $\mathbb{E}\left[\boldsymbol\Sigma^{-1}_{[\mathbf{Z}\mathbf{Z}]}\right] = \left(\nu - |\mathbf{Y}|\right)\boldsymbol\Psi^{-1}_{[\mathbf{Z}\mathbf{Z}]}$
\end{proof}


Based on Property \ref{prop:EVGB} and given the quadratic form of Problem GB's objective function, once $\mathbb{E}\left[\mathbf{Q}\right]$ and $\mathbb{E}\left[\boldsymbol\Sigma^{-1}_{[\mathbf{Z}\mathbf{Z}]}\right]$ are calculated, their eigenvalues can be identified, and their convexity can be characterized according to the value of $w_1$. The proofs follow exactly the same form as in Properties \ref{propWB_concave} and \ref{propWB_convex}.
Likewise, bounds on the value of $u_1$ can be readily identified as in Equations \eqref{eqConcaveBound} and \eqref{eqConvexBound}.

Even with a normal-inverse-Wishart prior, the parameters $\mathbf{v}$ and $\boldsymbol\Sigma^{-1}_{[\mathbf{Z}\mathbf{Z}]}\boldsymbol\mu_{[\mathbf{Z}]}$ do not have conveniently defined distribution under Problem GB and, as such, lack a conveniently defined expectation. Although the normal-inverse-Wishart prior allows the attacker to derive insights about how Problem GB behaves, finding an optimal solution to the problem remains challenging. The same issue arises when using alternative priors as well. Therefore, given this collective difficulty, Problem GB cannot be solved in the same manner as Problem WB. However, as discussed by \citet{powell2019unified}, there do exist methods to approximate optimal solutions under such conditions. We explore a few such approaches in the next section.

}

\subsection{Solution Techniques for the Grey-box Setting} \label{sec:SolvingGB}

Based on the work of \citet{powell2019unified}, we discuss herein two techniques to solve Problem GB with demonstrated empirical success. First, we present an adaption of Sample Average Approximation (SAA), a canonical stochastic programming method. Subsequently, we discuss the adaptation of stochastic-search methods for Problem GB. 


\subsubsection{Sample Average Approximation}

Rather than optimizing $\mathbb{E}\left[w_1 \phi_1(\mathbf{z}) +  w_2 \phi_2(\mathbf{z})\right]$, $\mathbf{z}^*$ can be estimated with the sample average approximation (SAA) method. This requires generating samples $j=1,...,J$ of $(\boldsymbol\mu, \boldsymbol\Sigma)$ and calculating the resulting parameters, e.g., $\mathbf{Q}_j$. The average of these randomly sampled functions is then utilized to estimate  $\mathbb{E}\left[w_1 \phi_1(\mathbf{z}) +  w_2 \phi_2(\mathbf{z})\right]$ yielding the following approximation to Problem GB:

\begin{align*}
\textbf{GB2}: \ &\max_{\mathbf{z} \in \mathcal{Z}} \ \frac{1}{J} \sum_{j=1}^J w_1 \phi_{1,j}(\mathbf{z}) +  w_2 \phi_{2,j}(\mathbf{z}).
\end{align*}

\noindent While not exact, Problem GB2 provides a tractable means of approximating Problem GB and, by the Law of Large Numbers,  the approximate objective function converges toward the true expectation as $J\to\infty$. Moreover, after sampling $J$ parameters, an optimal solution to Problem GB2 coincides with an optimal solution of

\begin{align*}
\max_{\mathbf{z} \in \mathcal{Z}} \  \mathbf{z}^T\left( \frac{1}{J} \sum_{j=1}^J w_1\mathbf{Q}_j- w_2 \boldsymbol\Sigma^{-1}_{[\mathbf{Z}\mathbf{Z}],j}\right) \mathbf{z} + \mathbf{z}^T\left(\frac{1}{J} \sum_{j=1}^J w_1 \mathbf{v}_j + 2w_2\boldsymbol\Sigma^{-1}_{[\mathbf{Z}\mathbf{Z}],j}\boldsymbol\mu_{[\mathbf{Z}], j} \right)
\end{align*}

\noindent implying that, after the requisite pre-processing, an optimal solution to Problem GB2 can also be found efficiently utilizing the same techniques discussed in Section \ref{sec:SolvingWB}. 

\subsubsection{Stochastic Gradient Ascent Techniques}

Whereas SAA sequentially samples and then solves an approximation of Problem GB, stochastic gradient ascent (SGA) methods iteratively sample parameters and use them to identify a new solution. Noting that 


\begin{align*}
\nabla \Big(w_1\phi_1(\mathbf{z}) +w_2\phi_2(\mathbf{z}) \Big) &= 2\Big( w_1 \mathbf{Q}-  w_2 \boldsymbol\Sigma^{-1}_{[\mathbf{Z}\mathbf{Z}]}\Big)\mathbf{z} + w_1 \mathbf{v} + 2w_2\boldsymbol\Sigma^{-1}_{[\mathbf{Z}\mathbf{Z}]}
\boldsymbol\mu_{[\mathbf{Z}]}, 
\end{align*}

\noindent a basic SGA approach samples $\{ \mathbf{Q}_j, \boldsymbol\Sigma_{[\mathbf{Z}\mathbf{Z}],j}^{-1}, \mathbf{v}, \boldsymbol\mu_{[\mathbf{Z}], j}\}$, and finds a new candidate attack via

$$
\mathbf{z}_{j+1} = \mathbf{z}_{j} + \alpha\mathbf{d}_j
$$

\noindent wherein $\alpha$ is a step size, and $\mathbf{d}_j$ is a unit vector given by


\begin{align*}
\mathbf{r}_j &= 2\Big( w_1 \mathbf{Q}_j-  w_2 \boldsymbol\Sigma^{-1}_{[\mathbf{Z}\mathbf{Z}]}\Big)\mathbf{z}  + w_1 \mathbf{v} + 2w_2\boldsymbol\Sigma^{-1}_{[\mathbf{Z}\mathbf{Z}]}\boldsymbol\mu_{[\mathbf{Z}] \boldsymbol\mu_{[\mathbf{Z}]}}, \\
\mathbf{d}_j &= \frac{1}{|| \mathbf{r}_j||} \mathbf{r}_j.
\end{align*}

\noindent The basic SGA method repeats this sampling and search until some convergence criteria is met, e.g., $|\mathbf{z}_{j+1}-\mathbf{z}_j|\le \varepsilon$ for some $\varepsilon>0$. Since Problem GB is a constrained optimization problem, the direction $\mathbf{d}_j$  is deflected when necessary to ensure feasibility. 

Contemporary adaptations to the basic SGA method are also applicable. We focus on three such methods used to great success in deep learning: AdaGrad, RMSProp, and Adam. Each of these methods are conceptually similar to the basic SGA approach; however, $\alpha$ is determined separately for each iteration-and-dimension pair based on a static learning rate and accumulations of previously observed gradients and in, the case of Adam, $\mathbf{d}_j$ is a linear combination of all past gradients. For effective usage, each of these methods must be tuned based on their associated hyperparameters.

More specifically, AdaGrad modifies the standard SGA approach by re-interpreting $\alpha$ as a distinct hyperparameter and using it to determine the step size based on the squares of previously evaluated gradients. Defining $\mathbf{R}_j = \sum_{k=1}^{j} \mathbf{r}_k \mathbf{r}_k^T$, AdaGrad's update rule is 

$$
\mathbf{z}_{j+1} = \mathbf{z}_{j} +  \text{diag}(\mathbf{R}_j+\varepsilon \mathbf{I})^{-\frac{1}{2}} \odot \alpha\mathbf{d}_j
$$

\noindent where $\varepsilon$ is some small real number, $\text{diag}(\cdot)$ creates of a vector of its arguments diagonal elements, $\odot$ is the Hadamard product, and the exponent is performed element-wise. The RMSProp update rule is the same as AdaGrad; however, it modifies its calculation of $\mathbf{R}_j$. Namely, $\mathbf{R}_j$ is a weighted average of past outer products such that $\mathbf{R}_j = \tau_1 \mathbf{R}_{j-1} + (1-\tau_1) \mathbf{r}_j \mathbf{r}_j^T $ where $\tau_1 \in (0,1)$ is the decay rate. Adam further extends RMSProp by incorporating momentum and a bias correction step. That is, $\mathbf{m}_j =\tau_2 \mathbf{m}_{j-1} + (1-\tau_2) \mathbf{r}_j $ is the momentum vector where $\tau_2 \in (0,1)$ is a decay hyperparameter; the bias corrected momentum is $\mathbf{\hat{m}}_j = (1-\tau_2^j)^{-1}\mathbf{m}_j$. Introducing an additional bias correction to the $\mathbf{R}_j$ from RMSProp , i.e., $\mathbf{\hat{R}}_j = (1-\tau_1^j)^{-1} \mathbf{R}_j$, Adam's update rule is defined as

$$
\mathbf{z}_{j+1} = \mathbf{z}_{j} +  \text{diag}(\mathbf{\hat{R}}_j+\varepsilon \mathbf{I})^{-\frac{1}{2}}\odot \alpha\mathbf{\hat{m}}_j.
$$

\noindent We refer the reader to \citet{AdaGrad_Duchi}, \citet{RMSProp_Tielman}, and \citet{Adam_Kingma}, respectively, for additional information on these techniques.

\section{Applications} \label{sec5}

We demonstrate the efficacy of our attack on three applications that focus on corrupting an explicit MVG model, a linear regression model, and a linear Gaussian state space model, respectively. The scenarios underlying each are motivated by \citet{valentini2013modeling}, \citet{aktekin2013assessment}, and \citet{luo2021multiple}. In each applied example, we explore the utility of the designed attacks in a tractable manner; as such, they are deliberately simple. Reality is often more complicated, but we desire the clearest example that nonetheless exercises all the machinery and modeling choices introduced.

Herein, we demonstrate algorithmic performance, showcase varied settings in which the attacks are applicable, and explore the impacts our methods have in each application. Attacks are benchmarked against $\mathbf{z}'$ as well as a random noise (RN) attack that forms $\mathbf{z}$ by adding uniform random noise to $\mathbf{z}'$ while ensuring feasibility. To reproduce these examples, please find publicly available code online at https://github.com/mrlarosa/DistruptingMVGs. 

\subsection{Attacking an Explicit Multivariate Gaussian Model}

Since 2000, Zillow has reported the Zillow home value index (ZHVI) for single-family homes on a subset of Arizona counties \citep{Zillow}; however, only since 2016 has the ZHVI included all of its 15 counties. We consider an attacker corrupting a decisionmaker's conditional MVG on ZHVI-values for the unobserved counties in January 2000 used for, e.g., tax assessment decisions. In January 2000, ZHVI-values were calculated for all counties except Yuma ($Y_1$), Cochise ($Y_2$), Apache ($Y_3$), and La Paz ($Y_4$). The observed ZHVI values for Maricopa ($Z_1$), Pima ($Z_2$), Pinal ($Z_3$), Yavapai ($Z_4$), Mohave ($Z_5$), Coconino ($Z_6$), Navajo \textbf{($Z_7$)}, Gila ($Z_8$), Santa Cruz ($Z_9$), Graham ($Z_{10}$), and Greenlee ($Z_{11}$) counties in hundreds of thousands of dollars are 1.459, 1.214, 1.344, 1.377, 0.941, 1.531, 0.838, 0.766, 0.679,  0.856, 0.537, respectively. The attacker does not wish to modify any observation by more than fifteen thousand dollars (i.e., 0.15) to ensure the corrupted observation remains credible. Under base conditions, we assume $u_1=u_2=0.5$. We explore each attack to gain insight and characterize their implications. Results are summarized in Table \ref{tab: Zillow} for ease of reference.

\begin{table}[htbp!]
\centering
\caption{Benchmarking attacks on the ZHVI Problem} \label{tab: Zillow}
\resizebox{140mm}{!}{
\begin{tabular}{ccccccccccccccccc}
\hline 
Attack & Variant  & Hyperparameters$^\dag$     & $z_1$ & $z_2$ & $z_3$ & $z_4$ & $z_5$ & $z_6$  & $z_7$ & $z_8$ & $z_9$ & $z_{10}$ & $z_{11}$ & Obj. Value$^\ddag$ & $D_{KL}$$^{**}$ & Comp. Effort (sec)  \\ \hline
WB & - & - & 1.609 & 1.065 & 1.194 & 1.527 & 0.862 & 1.681 & 0.989 & 0.917 & 0.807 & 0.707 & 0.513 & -0.1142 & 255.308 & 0.679 \\ \cdashline{1-17} 
\multirow{5}*{SAA}& \multirow{5}*{-} & 25&	1.609&	1.065&	1.194&	1.527&	0.858&	1.681&	0.989&	0.917&	0.787&	0.707&	0.546&	-0.128&	256.307 & 0.381 \\
& & 100&	1.609&	1.065&	1.194&	1.527&	0.863&	1.681&	0.989&	0.917&	0.789&	0.707&	0.485&	-0.162 & 261.249 &	0.541 \\
& & 500&	1.609&	1.065&	1.194&	1.527&	0.848&	1.681&	0.989&	0.917&	0.794&	0.707&	0.499&	-0.115 & 260.796 &	0.977 \\
& & 2500&	1.609&	1.065&	1.194&	1.527&	0.854&	1.681&	0.989&	0.917&	0.798&	0.707&	0.520&	-0.129& 257.206 &	2.509 \\
& & 10000&	1.609&	1.065&	1.194&	1.527&	0.856&	1.681&	0.989&	0.917&	0.801&	0.707&	0.517&	-0.134& 256.793 &	9.876  \\ \cdashline{1-17} 

\multirow{9}{*}{SGA$^*$} & \multirow{3}{*}{AdaGrad} & $\{0.01, 10^{-8}\}$ &  1.609&	1.065&	1.194 &	1.527 &	0.851 &	1.681 &	0.989 &	0.917 &	0.798 &	0.707 &	0.518 &	-0.1343 & 257.794 & 27.470 \\
& & $\{0.025, 10^{-7}\}$ &1.609 &	1.065 &	1.194 &	1.526 &	0.854 &	1.681 &	0.989 &	0.917 &	0.803 &	0.707 &	0.508 &	-0.1342& 257.503 & 	27.066 \\
& & $\{0.05, 10^{-6}\}$ &1.609 &	1.065 &	1.194 &	1.527 &	0.853 &	1.681 &	0.989 &	0.917 &	0.806 &	0.707 &	0.508 &	-0.1394& 257.091 & 	63.873 \\ \cdashline{2-17}
 & \multirow{3}{*}{RMSProp} & $\{0.001, 10^{-5}, 0.9\}$ &1.609 &	1.065 &	1.194 &	1.527 &	0.853 &	1.681 &	0.989 &	0.917 &	0.803 &	0.707 &	0.507 &	-0.1343& 257.722 & 	49.004 \\
& &  $\{0.0005, 10^{-6}, 0.85\}$ &1.544 &	1.065&	1.194 &	1.527 &	0.814&	1.681 &	0.886 &	0.917 &	0.690 &	0.707 &	0.489 &	-0.1465& 205.551 & 	43.173 \\
& & $\{0.0001, 10^{-7}, 0.8\}$ &1.309 &	1.065 &	1.194 &	1.228 &	0.792 &	1.382&	0.689 &	0.617 &	0.530 &	0.707 &	0.388 &	-0.4284& 3.8644 & 	13.294 \\ \cdashline{2-17}
 & \multirow{3}{*}{Adam} & $\{0.001, 10^{-8}, 0.9, 0.9\}$ &1.607 &	1.065 &	1.194 &	1.527 &	0.850 &	1.681 &	0.988 &	0.917 &	0.790 &	0.707&	0.531 &	-0.1326& 256.906 & 	72.767 \\
& & $\{0.0025, 10^{-7}, 0.8, 0.8\}$ &1.609 &	1.065 &	1.194 &	1.527 &	0.859 &	1.681 &	0.989 &	0.917 &	0.770 &	0.707 &	0.539 &	-0.1344& 260.205 & 	70.839 \\
& & $\{0.005, 10^{-6}, 0.7, 0.7\}$ &1.609 &	1.065&	1.194&	1.527 &	0.855 &	1.681&	0.989 &	0.917&	0.782&	0.707&	0.535&	-0.1348& 258.817 & 	81.581 \\ \cdashline{1-17} RN & - & - & 1.494 & 1.073 & 1.241 & 1.431 & 0.920 & 1.580 & 0.959 & 0.880 & 0.738 & 0.798 & 0.427 & -0.233 & 108.998 & 0.069 \\ \cdashline{1-17} 
$\mathbf{z}'$ & - &- & 1.459 & 1.214 & 1.344 & 1.377 & 0.941 & 1.531 & 0.838 & 0.766& 0.679 & 0.856 & 0.537 & -0.427 & 0 & - \\ \hline 
\multicolumn{15}{l}{$^\dag$ SAA hyperparameter is $J$; AdaGrad is $\{\alpha, \varepsilon\}$, RMSProp is $\{\alpha, \varepsilon, \tau_1\}$, and Adam is $\{\alpha, \varepsilon, \tau_1, \tau_2\}$}\\ 
\multicolumn{15}{l}{$^\ddag$ WB, RN and $\mathbf{z}'$ values correspond to whitebox objective; SAA and SGA to grey-box}\\ 
\multicolumn{15}{l}{$^*$ Objective function values approximated via 10,000 Monte Carlo simulations of the indicated attack }\\
\multicolumn{15}{l}{$^{**}$ KL divergence between induced and true conditional treating the white-box model as the true joint}
\end{tabular}
}
\end{table}

\subsubsection{White-box Setting} \label{sec:WBZillow}

Assume the decisionmaker fits an MVG to the ZHVI data using the county scores recorded monthly from May 2016 to February 2023. This joint distribution over ZHVI values describes a fifteen dimensional random variable that is known to the attacker under white-box conditions. The exact values of the $\boldsymbol\mu$ and $\boldsymbol\Sigma$ parameters are available online in our code repository.

Bounds on the convexity of Problem WB's objective function are calculated as described in Section \ref{sec:SolveWB}. It is found that $u_1^- = 0.000037$ and $u_1^+ = 1$ implying that, under the base conditions, Problem WB's objective function is potentially neither convex nor concave, and typical quadratic programming methods are likely inapplicable. However, leveraging CPLEX, we approximate the optimal attack as depicted in Table \ref{tab: Zillow} with an objective function value of -0.1142 in less than a second of effort. The KL divergence between the true and corrupted conditional distributions is 255.308.

\begin{figure}[htbp!]
    \centering
    \includegraphics[width=50mm]{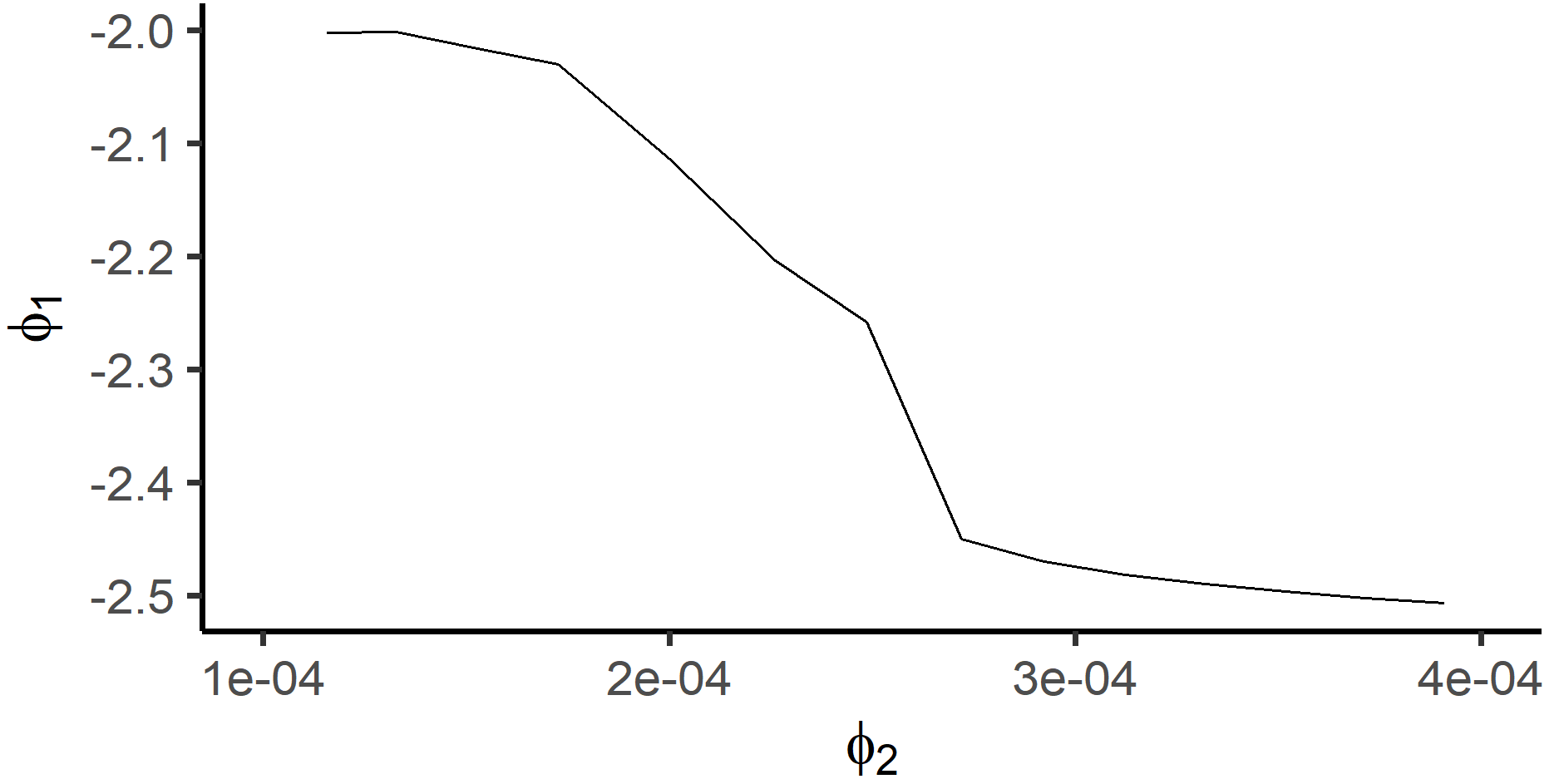}
    \caption{Approximate Pareto front for the white-box ZHVI Problem}
    \label{fig:Pareto}
\end{figure}

Based on the multi-objective nature of the problem, additional analysis may estimate a set of non-dominated solutions under all $u_i$-values. We approximate this Pareto front in Figure \ref{fig:Pareto} by solving Problem WB under $u_1 \in \{0.01, 0.05, 0.1,..., 0.95, 0.99\}$ and $u_2 = 1-u_1$. Based on the values of $u_1^-$ and $u_1^+$, the objective function is potentially neither convex-nor-concave in much of this region. We leverage algorithmic settings in CPLEX to accommodate this difficulty but, since the solutions are not assuredly optimal, Figure \ref{fig:Pareto} displays an approximation. 


\subsubsection{Grey-box Setting} \label{sec:GBZillow}

In the grey-box setting, we assume the attacker does not know the decisionmaker's exact MVG but has some belief about its parameterization. The attacker models their belief about this distribution via a normal-inverse-Wishart prior such that $\boldsymbol\mu_0 = \hat{\boldsymbol\mu}$, $\kappa =5$, $\boldsymbol\Psi = \hat{\boldsymbol\Sigma}$, and $\nu = 17$ wherein $\hat{\boldsymbol\mu}$ and $\hat{\boldsymbol\Sigma}$ are the maximum likelihood estimates from the previous section. This parameterization assures that the attacker's belief about the decisionmaker's MVG are true in expectation. To ensure proper comparison, we use the same $\phi_1^*$ and $\phi_2^*$ as identified via SAA.

Table \ref{tab: Zillow} displays the corrupted data and objective-function values identified by the attacker utilizing the SAA method with varied $J$. We also include the computational effort required to generate the sample and solve the instance, the KL divergence to the decisionmaker's true conditional distribution (defined using the white-box model), and the objective-function value. As expected, the solutions converge as $J$ increases. The estimate of the objective function value improves as well. For larger $J$, additional computational effort is required but, even for $J=10000$, our approach solves the problem instance in under ten seconds.  

In juxtaposition, Table \ref{tab: Zillow}  also displays the corrupted data, objective-function value, KL divergence, and required computational effort for the SGA methods under varied hyperparameterizations. Algorithmic termination occurs when the solution moves less than 0.0001 units in Euclidean distance. For AdaGrad, RMSProp and Adam, the set of hyperparameters respectively represent $\{\alpha, \varepsilon\}$, $\{\alpha, \varepsilon, \tau_1\}$. and $\{\alpha, \varepsilon, \tau_1, \tau_2\}$. Table \ref{tab: Zillow} shows that each of these methods tend to converge near the SAA-solution, but required more computational effort.

\subsubsection{Implications and Insights}
Each attack achieves its goal of indiscriminately disrupting inference while maintaining the plausibility of $\mathbf{z}$. We discuss herein these implications globally and on a specific estimation task. Likewise, we gain insight by comparing and contrasting the attacks against each other as well as the RN baseline. 

As judged by the KL divergence to the true conditional distribution, the white-box attack successfully modifies the decisionmaker's model and outperforms the RN baseline, thereby indiscriminately corrupting inference. Should we also consider the decisionmaker's modal estimate of $\mathbf{Y}|\mathbf{z}$ as a concrete example, the effects become even more apparent. Under $\mathbf{z}'$ the modal estimate is (0.98, 0.45, 1.09, 0.41) but, under $\mathbf{z}$, it is (0.88, 0.75, 0.77, 0.48); more drastic effects can be induced by enlarging $\mathbf{Z}$ or increasing $u_1$. Inspection of the white-box attack yield other insights as well. It generally pushes $z_i$ to the edge of the feasible region; however, this did not occur for $z_{11}$ whose value was selected within the interior of $\mathcal{Z}$. In juxtaposition to the fast-gradient-sign method \citep{goodfellow2014explaining} that selects $\mathbf{z}$ on the edge of some $\varepsilon$-neighborhood about $\mathbf{z}'$, our attacks more readily search the interior of $\mathcal{Z}$. Yet, the behavior of $z_{11}$ in this white-box instance leads one to question when such behavior may occur. Intuitively, if some $z'_i$ is near its conditional mode and $Z_i$ is not associated with any other $Z_j$ or $\mathbf{Y}$, its value does not affect the KL divergence yet its modification incurs a penalty. In general, the association need not be zero; however, its effect must be overwhelmed accordingly by the objective-function weights.

Assuming the white-box model is the decisionmaker's true distribution, we see that both grey-box attacks induce large KL divergences as well. Every SAA attack and all but one SGA attack outperform the RN baseline. As may be expected, this suggests the SGA attacks require more tuning; however, since the SAA attacks perform well with low $J$, they show themselves to be an effective heuristic even under a limited computational budget. For the specific modal estimation task, the SAA attack with $J=10000$ induced an estimate of (0.89, 0.75, 0.77, 0.48); the relatively unsuccessful SGA attack using RMSProp with $\{ 0.001, 10^{-7}, 0.8\}$ also yielded an estimate of (0.92, 0.36, 0.93, 0.30). Moreover, based on Property \ref{prop:EVGB}, we see that the white- and grey-box objective functions are distinct, even if the attacker's beliefs are true in expectation (e.g., $\mathbb{E}[\mathbf{Q}] \ne \mathbf{Q}$). This means that, despite the attacker's relatively accurate beliefs, the two problems may have disparate solutions. However, in this setting, we see that the geometry of each problem, when scaled by their respective $\phi_i^*$, is relatively similar. This induces comparable $\mathbf{z}$- and objective function values. 

Nevertheless, as judged by the KL divergence, we can see that the grey-box attacks tend to be slightly more aggressive. Further exploration reveals, however, that this behavior is an artifact of the prior. Utilizing the SAA method with $J= 1000$, we examine the effect that alternative priors have on the attacker's action. Table \ref{tab: AZ_GB_prior} provides the identified attack using varying priors. Since their objective functions are distinct, these priors induce diverse behavior as evidence by their $\mathbf{z}$-values and KL divergence to the true conditional. The table makes clear the effect of the attacker's prior; if their beliefs are inaccurate so too is the attack. 

\begin{table}[htbp!]
\caption{Evaluating SAA attacks on the ZHVI Problem using alternative priors} \label{tab: AZ_GB_prior}
\centering
\resizebox{112mm}{!}{
\begin{tabular}{ccccccccccccccc}
\hline 
Attack & $\left(\boldsymbol\mu_0, \kappa, \boldsymbol\Psi\ , \nu \right)$      & $z_1$ & $z_2$ & $z_3$ & $z_4$ & $z_5$ & $z_6$  & $z_7$ & $z_8$ & $z_9$ & $z_{10}$ & $z_{11}$ & Obj. Value$^\ddag$ & $D_{KL}$ $^{**}$ \\ \hline
\multirow{4}{*}{SAA}  & $\left(\hat{\boldsymbol\mu}, 5, 2\hat{\boldsymbol\Sigma} + \mathbf{I}, 18 \right)$&	1.609&	1.301&	1.194&	1.527&	1.092&	1.681&	0.989&	0.917&	0.829&	1.007&	0.604 & 0.0013 & 58.377\\
& $\left(\hat{\boldsymbol\mu}, 10, 3\hat{\boldsymbol\Sigma} , 19 \right)$&	1.609&	1.065&	1.194&	1.527&	0.858&	1.681&	0.989&	0.917&	0.801&	0.707&	0.506 & -0.0429 & 257.574\\
& $\left(0.5\hat{\boldsymbol\mu}, 1, 0.5\hat{\boldsymbol\Sigma} , 18 \right)$&	1.609&	1.065&	1.194&	1.246&	0.841&	1.681&	0.989&	0.917&	0.822&	0.707&	0.388 & -0.9962 & 309.905\\
& $\left(2\hat{\boldsymbol\mu}, 10, 2\hat{\boldsymbol\Sigma} , 36 \right)$&	1.417&	1.365&	1.194&	1.527&	0.792&	1.681&	0.938&	0.877&	0.829&	0.707&	0.688 & 1.457 & 38.4578 \\ \cdashline{1-15}
RN & - & 1.494 & 1.073 & 1.241 & 1.431 & 0.920 & 1.580 & 0.959 & 0.880 & 0.738 & 0.798 & 0.427 & -0.233 & 108.998 \\  \cdashline{1-15} $\mathbf{z}'$ &-  & 1.459 & 1.214 & 1.344 & 1.377 & 0.941 & 1.531 & 0.838 & 0.766& 0.679 & 0.856 & 0.537 & -0.427 & 0\\ \hline
\multicolumn{15}{l}{$^\ddag$ WB, RN and $\mathbf{z}'$ values correspond to whitebox objective; SAA and SGA to grey-box}\\ 
\multicolumn{15}{l}{$^{**}$ KL divergence between induced and true conditional treating the white-box model as the true joint}
\end{tabular}
}
\end{table}


Finally, we note that, for illustrative purposes, we assumed the decisionmaker fit his MVG upon the Arizona ZHVI-values using traditional maximum likelihood estimation. This implies the decisionmaker treats the observations as independent and identically distributed. Such treatment is likely ill-advised given the temporal nature of their data; however, we highlight that the attacker's effect depends upon the accuracy of their beliefs, not necessarily on the validity of the decisionmaker's model in and of itself. 

\subsection{Attacking a Linear Regression Model}

A decisionmaker desires to estimate the interest rate of a loan ($Y$) based on a set of predictors associated with the individual who receives it. These predictors include their income ($Z_1$), their debt-to-income ratio ($Z_2$), their total credit balance ($Z_3$), their total debit limit ($Z_4$), their total credit limit ($Z_5$), the percentage of credit lines without a delinquency ($Z_6$), and the loan amount ($Z_7$). In this example, we use a subset of loan data from Lending Club, available for download from \citet{OpenIntro2023}. All monetary values are represented in terms of thousands of dollars. 

Formally, we assume that the decisionmaker has fit a linear regression model mapping the conditional distribution of $Y$ given the predictors as 

\begin{align*}
Y| \mathbf{z} \sim \mathcal{N}(\beta_0 +\boldsymbol\beta^T \mathbf{z}, \sigma^2).
\end{align*}

\noindent Likewise, we augment this standard regression model by assuming that, for an arbitrary individual, the decisionmaker believes the predictors have a MVG distribution. 

This section explores how our attacks can be used to corrupt the decisionmaker's conditional beliefs about $Y|\mathbf{z}$ in the white- and grey-box settings based on the results presented by \citet{koller2009probabilistic}. \citet{koller2009probabilistic} show that, in a Gaussian Bayesian network, if a variable is a linear Gaussian of its parents, the joint distribution of the variable and its parents is also Gaussian. Therefore, since we have $\mathbf{Z} \sim \mathcal{N}(\boldsymbol\mu_{[\mathbf{Z}]}, \boldsymbol\Sigma_{[\mathbf{Z}\mathbf{Z}]})$ and $Y| \mathbf{z} \sim \mathcal{N}(\beta_0 +\boldsymbol\beta^T \mathbf{z}, \sigma^2)$, their results imply $(\mathbf{Z},Y)\sim \mathcal{N}(\boldsymbol\mu, \boldsymbol\Sigma)$ such that 

$$ 
\boldsymbol\mu = \begin{bmatrix} \boldsymbol\mu_{[\mathbf{Z}]} \\ \boldsymbol\mu_{[Y]} \end{bmatrix}, \quad 
\boldsymbol\Sigma = \begin{bmatrix} \boldsymbol\Sigma_{[\mathbf{Z}\mathbf{Z}]} & \boldsymbol\Sigma_{[\mathbf{Z}Y]}\\
\boldsymbol\Sigma_{[Y\mathbf{Z}]} & \boldsymbol\Sigma_{[YY]} \end{bmatrix},
$$

\noindent where

\begin{align*}
\boldsymbol\mu_{[Y]} &= \beta_0 + \boldsymbol\beta^T \boldsymbol\mu_{[\mathbf{Z}]} ,\\
\boldsymbol\Sigma_{[Y Z_i]} &= \boldsymbol\beta^T \boldsymbol\Sigma_{[\mathbf{Z} Z_i]}, \ \forall Z_i, \\ 
\boldsymbol\Sigma_{[Y Z_i]} &= \boldsymbol\Sigma_{[Z_i Y]}, \ \forall Z_i, \\
\boldsymbol\Sigma_{[Y Y]} &= \sigma^2 + \boldsymbol\beta^T \boldsymbol\Sigma_{[\mathbf{Z}\mathbf{Z}]} \boldsymbol\beta
\end{align*}

\noindent This representation enables the direct utilization of our attacks. 

In what follows, we assume the attacker wishes to corrupt the decisionmaker's conditional distribution about the interest rate to be received by a specific individual. This is accomplished by corrupting 
$\mathbf{z}'=$(90.000, 18.010, 38.767, 11.100, 70.795, 28.000, 92.900). 
To avoid immediate detection, the attacker is unwilling to corrupt any data point by more than 10\% of its true value. Under base conditions, we again assume $u_1=u_2=0.5$. Results are summarized in Table \ref{tab: Loan_GB_J} for reference. 

\begin{table}[htbp!]
\centering
\caption{Benchmarking attacks on the Loan Problem} \label{tab: Loan_GB_J}
\resizebox{112mm}{!}{
\begin{tabular}{cccccccccccccccc}
\hline
Attack & $J$     & $z_1$ & $z_2$ & $z_3$ & $z_4$ & $z_5$ & $z_6$  & $z_7$ & Obj. Value$^\ddag$ & $D_{KL}$$^{**}$ & Comp. Effort (sec)  \\ \hline
WB & - & 81.000 & 16.209 & 38.272 & 12.210 & 77.874 & 25.200 & 102.190 & -0.004 & 0.007 & 0.124\\ \cdashline{1-12}
\multirow{5}*{SAA}& 25&	81.000&	17.759&	34.890&	12.210&	77.874&	25.200&	91.706&	0.09504 & 5.67e-5 & 0.127 \\
& 100&	81.000&	16.352&	34.890&	12.210&	77.875&	25.200&	95.624&	0.0501&	 7.91e-5 & 0.226 \\
& 500&	81.000&	16.209&	34.890&	12.210&	77.874&	25.200&	94.742&	0.0323&	6.88e-4 & 0.873 \\
& 2500&	81.000&	16.608&	42.644&	12.210&	77.874&	25.200&	94.657&	0.0476&	9.32e-6 &  4.278\\
& 10000&	81.000&	16.209&	34.890&	12.210&	77.875&	25.200&	94.389&	0.0555&	5.52e-4 & 21.031 \\ \cdashline{1-12} RN & - & 92.107 & 16.310 & 36.101 & 11.499 & 69.789 & 28.911 & 100.331 & -0.011& 0.0047 & 0.031 \\ \cdashline{1-12} $\mathbf{z}'$& - & 90.000 & 18.010 & 38.767 & 11.100 & 70.795 & 28.000 & 92.900  & -0.0144  & 0  & -\\\hline
\multicolumn{15}{l}{$^\ddag$ WB, RN and $\mathbf{z}'$ values correspond to whitebox objective; SAA and SGA to grey-box}\\ 
\multicolumn{15}{l}{$^{**}$ KL divergence between induced and true conditional treating the white-box model as the true joint}
\end{tabular}
}
\end{table}

\subsubsection{White-box Setting}

In the white-box setting, we assume the decisionmaker has fit a MVG to the predictor variables, and separately has fit a linear regression model with standard techniques. This is accomplished using all but the first observation provided by \citet{OpenIntro2023}; the first observation serves is $\mathbf{z}'$ as previously defined. Akin to Section \ref{sec:WBZillow}, the identified parameterization is available online in our code repository. 

\begin{figure}[htbp!]
    \centering
    \includegraphics[width=45mm]{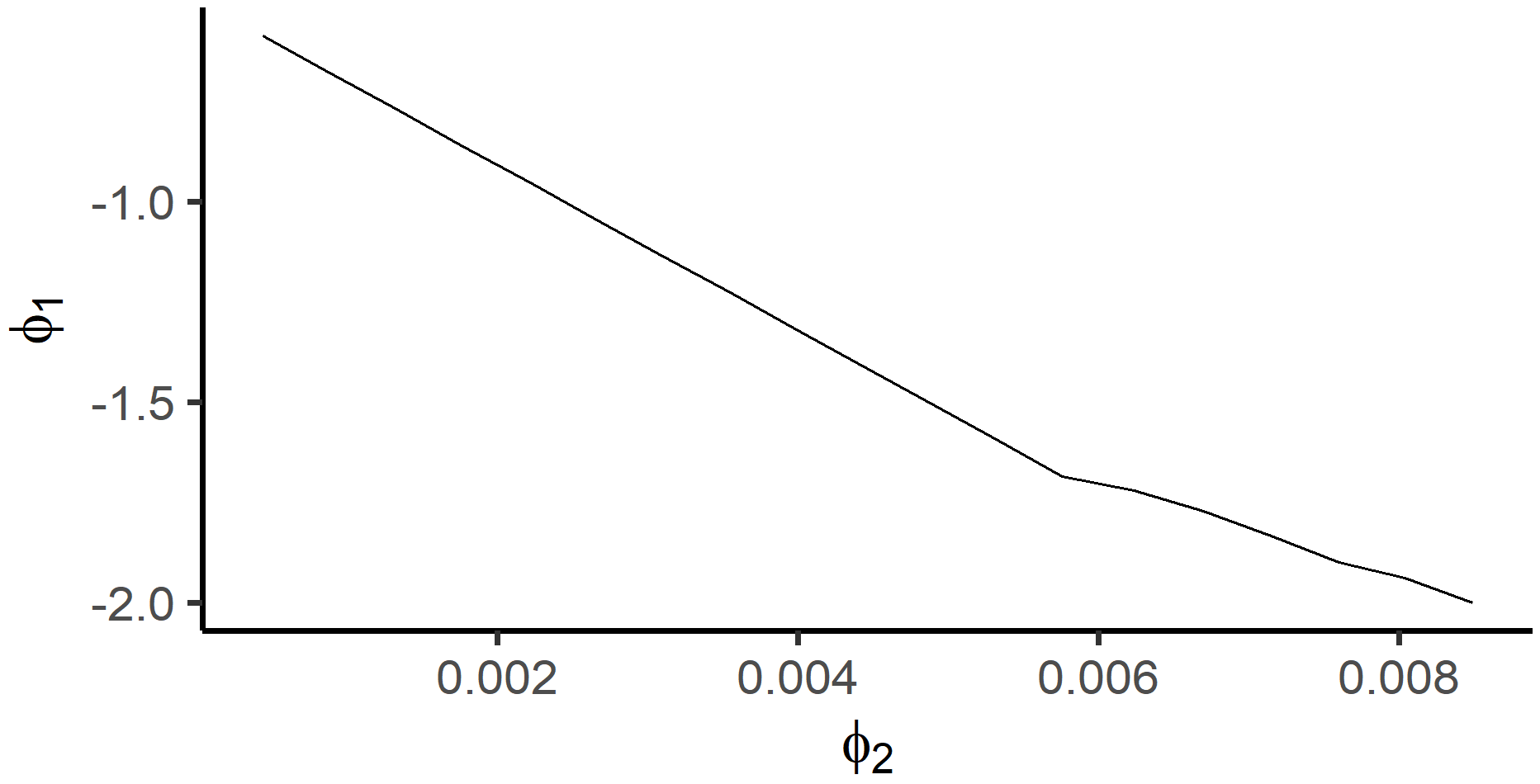}
    \caption{Approximate Pareto front for the white-box Loan Problem}
    \label{fig:Pareto2}
\end{figure}

Utilizing the results of Section \ref{sec:SolveWB}, we identify $u_1^- = 0.0008$ and $u_1^+ = 1$, implying that, under base conditions, the attacker may be solving a non-convex quadratic program; this is verified analytically to be the case. 
Utilizing CPLEX and an appropriate solution method, we approximate the optimal attack
$\mathbf{z} $ = (81.000, 16.209,	38.272,	12.210, 77.874, 25.200, 102.190) 
with an objective function value of  $-0.004$. The KL divergence to the true conditional is 0.007. We provide the approximated Pareto front for this problem instance in Figure \ref{fig:Pareto2} whereby we see a similar pattern to that discussed in Figure \ref{fig:Pareto}. 

\subsubsection{Grey-box Setting} 

Although the attacker does not know the decisionmaker's model, we assume he models his beliefs using classical distributions. Namely, beliefs about $\mathcal{P}_\mathbf{Z}$ is normal-inverse-Wishart with $\boldsymbol\mu_0 = \hat{\boldsymbol\mu}_{[\mathbf{Z}]}$, $\kappa =5$, $\boldsymbol\Psi = \hat{\boldsymbol\Sigma}_{[\mathbf{Z}\mathbf{Z}]}$, and $\nu = 9$ where $\hat{\boldsymbol\mu}_{[\mathbf{Z}]}$ and $\hat{\boldsymbol\Sigma}_{[\mathbf{Z}\mathbf{Z}]}$ are the decisionmaker's true model parameters. This is independent of their beliefs about $\beta_0$, $\boldsymbol\beta$ and $\sigma^2$ whose joint distribution is modeled hierarchically. That is, $\sigma^2$ has an inverse-gamma distribution of $\mathcal{IG}(4,2)$ and $(\beta_0, \boldsymbol\beta)$ has a simple multivariate normal distribution centered on the true $(\beta_0, \boldsymbol\beta)$-values with a covariance matrix equal to $\sigma^2 \mathbf{I}$. 

Given these beliefs, samples of the decisionmaker's model and, by using the methods described by \citet{koller2009probabilistic}, in conjunction with the results of Property \ref{prop:KLinz}, the SAA attacks can be readily performed. We present results of this attack with varied values of $J$ in Table \ref{tab: Loan_GB_J}. Similar behavior is observed as in Table \ref{tab: Zillow}; computation time increases with $J$ but the solutions also converge. Likewise, some observations remain relatively static as $J$ is increased whereas others vary (e.g., $z_3$ and $z_7$). As with the ZHVI application, these SAA attacks are also identified expeditiously. 

\subsubsection{Implications and Insights}

The white- and grey-box attack effectively disrupt the decisionmaker's distribution; however, the feasible region and the nature of the model inhibits the degree to which this can occur. Assuming the white-box model is the decisionmaker's true model, the KL divergences between the true and adversarially induced conditional distributions is relatively small under both a white- and grey-box attack scheme. 
Consider the modal estimate of the individual's offered interest rate. Under $\mathbf{z}'$, its value is 12.7\% but, under the white-box attack, it is 12.1\%. 
Similar behavior is observed for the grey-box attacks; the SAA attack with $J=10000$ yield an estimated rate of 12.5\%.
Numerically, such values appear 
minuscule
however, small differences in interest rate over the life of a loan accumulate; such effects are even more drastic when multiple loans are targeted.

Note that, because there is a single latent variable in this specific setting, the indiscriminate disruption of the conditional distribution is solely concerned with the loan's interest rate.

Regarding baseline comparisons, all attacks outperform the RN baseline in terms of their objective-function value. The RN baseline identifies attacks that are more aggressive with respect to the KL divergence but do not adequately address the attacker's plausibility goal. Regarding computational effort, the RN baseline and SAA with $J=25$ have comparable demands; however, the SAA method finds a much improved solution. Moreover, akin to the previous example, it can be observed that the attacks found it beneficial to keep $z_7$ within the interior of $\mathcal{Z}$, highlighting the importance of systematic search methods. 

Alternatively, in juxtaposition to the previous example, the identified $\mathbf{z}$ and objective function values between the white- and grey-box attacks no longer coincide, despite the grey-box beliefs being centered upon the white-box model. This highlights that the geometry of the two problems may be distinct. Once more, we see the effects of Property \ref{prop:EVGB} in action. The attacker's expected value of the decisionmaker's model may correspond with truth, but that does not mean that the white- and grey-box objectives coincide. Such behavior highlights the benefits of complete knowledge from the attacker's perspective. For this specific example, it signifies that the attacker is well-served by locating an insider threat in the decisionmaker's financial institution (e.g., via coercion or other means). The decisionmaker must thus remain cognizant of this incentive structure and adopt appropriate operational-security strategies.

\subsection{Attacking a Linear Gaussian State Space Model}

Linear Gaussian state space models appear in a variety of engineering and signals-processing applications, and are useful for modelling other dynamic systems \citep[e.g., see][]{valentini2013modeling, mcdonald2021markov}. For example, they underpin the well-known Kalman filter and can be used to infer the true state of an object given noisy sensor measurements in discrete time. In this application, we consider a simplified LG-SSM that explores how our attacks can be utilized to corrupt inference about the movement of a physical object in a temporal setting.

More specifically, we consider an object moving at constant velocity in a two-dimensional environment whose state-transition model is given by 

\begin{align} \label{eqLGSSM1}
\small
\begin{bmatrix}
Y_{1,t}\\
Y_{2,t} \\
\dot{Y}_{1,t} \\
\dot{Y}_{2,t}
\end{bmatrix} = 
\begin{bmatrix} 
1 & 0& \Delta t &0 \\
0 & 1 &  0 & \Delta t\\
0 &0 & 1 & 0\\
0 & 0 & 0 & 1 \\
\end{bmatrix}
\begin{bmatrix}
Y_{1,t-1}\\
Y_{2,t-1} \\
\dot{Y}_{1,t-1} \\
\dot{Y}_{2,t-1}
\end{bmatrix} + \boldsymbol\varepsilon,
\end{align}

\noindent where $\boldsymbol\varepsilon \sim \mathcal{N}(\mathbf{0}, \mathbf{K}_{\boldsymbol\varepsilon })$ and $\mathbf{K}_{\boldsymbol\varepsilon } \in \mathbb{R}^{4\times 4}$. Variables $Y_{1,t}$ and $Y_{2,t}$ describe the observable true position of the object in dimension one and two, respectively; $\dot{Y}_{1,t}$ and $\dot{Y}_{2,t}$ correspond to its velocity in each dimension.  Moreover, the observation model is given by

\begin{align} \label{eqLGSSM2}
\small
\begin{bmatrix}
Z_{1,t}\\
Z_{2,t}
\end{bmatrix} = 
\begin{bmatrix} 
1 & 0& 0 &0 \\
0 & 1 &  0 & 0 \\
\end{bmatrix}
\begin{bmatrix}
Y_{1,t-1}\\
Y_{2,t-1} \\
\dot{Y}_{1,t-1} \\
\dot{Y}_{2,t-1}
\end{bmatrix}  + \boldsymbol\delta
\end{align}

\noindent where $\boldsymbol\delta \sim \mathcal{N}(\mathbf{0}, \mathbf{K}_{\boldsymbol\delta })$ and $\mathbf{K}_{\boldsymbol\delta} \in \mathbb{R}^{2\times 2}$.  Variables $Z_{1,t}$ and $Z_{2,t}$ are the observable sensor measurements on the object's position in the two-dimensional environment. Together, these equations define the decisionmaker's LG-SSM, and the decisionmaker uses the $Z$-values to make inference about the $Y$-values some finite set of times, $\mathcal{T}$. 

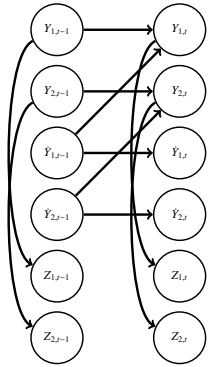
\begin{figure}[htbp!]
    \centering
\resizebox{28mm}{!}{
 \begin{tikzpicture}[->, 
roundnode/.style={circle, draw=black, very thick, minimum size=15mm},
]
\node[roundnode] (y11)            {$Y_{1,t-1}$};
\node[roundnode] (y21) [below=.25cm of y11] {$Y_{2,t-1}$};
\node[roundnode] (ydot11) [below=.25cm of y21] {$\dot{Y}_{1,t-1}$};
\node[roundnode] (ydot21) [below=.25cm of ydot11] {$\dot{Y}_{2,t-1}$};
\node[roundnode] (z11) [below=.25cm of ydot21] {$Z_{1,t-1}$};
\node[roundnode] (z21) [below=.25cm of z11] {$Z_{2,t-1}$};

\node[roundnode] (y12) [right=2cm of y11]            {$Y_{1,t}$};
\node[roundnode] (y22) [below=.25cm of y12] {$Y_{2,t}$};
\node[roundnode] (ydot12) [below=.25cm of y22] {$\dot{Y}_{1,t}$};
\node[roundnode] (ydot22) [below=.25cm of ydot12] {$\dot{Y}_{2,t}$};
\node[roundnode] (z12) [below=.25cm of ydot22] {$Z_{1,t}$};
\node[roundnode] (z22) [below=.25cm of z12] {$Z_{2,t}$};

\draw[->,line width=2pt] (y11) -- (y12);
\draw[->,line width=2pt] (y21) -- (y22);
\draw[->,line width=2pt] (ydot11) -- (ydot12);
\draw[->,line width=2pt] (ydot21) -- (ydot22);
\draw[->,line width=2pt] (ydot11) -- (y12);
\draw[->,line width=2pt] (ydot21) -- (y22);
\draw[->,line width=2pt] (y11) to [out = 205, in=155, looseness=0.4] (z11);
\draw[->,line width=2pt] (y12) to [out = 205, in=155, looseness=0.4] (z12);
\draw[->,line width=2pt] (y21) to [out = 205, in=155, looseness=0.4] (z21);
\draw[->,line width=2pt] (y22) to [out = 205, in=155, looseness=0.4] (z22);


\end{tikzpicture}
}
    \caption{A 2-TBN representation of the two-dimensional LG-SSM}
    \label{fig:2TBN}
\end{figure}

Instead of using Equations \eqref{eqLGSSM1} and \eqref{eqLGSSM2}, LG-SSMs are often parameterized as a Gaussian Bayesian network. Figure \ref{fig:2TBN} represents the decisionmaker's LG-SSM as a two-time-slice-Bayesian network (2TBN) that can be unrolled over $\mathcal{T}$ to represent the joint distribution over the random variables. The conditional distributions on the nodes are univariate Gaussians with a mean that is a linear combination of its parents and a variance that is shared across time for variables of the same type; the sole exception is the initial position which has its own probability distribution. By representing the LG-SSM in this manner, the recursion described by \citet{koller2009probabilistic} from the linear regression application defines the joint distribution. Moreover, as the physical model is known, the $\beta$-terms are given directly by Equations \eqref{eqLGSSM1} and \eqref{eqLGSSM2}, i.e., they equal one and $\Delta t$ where appropriate.

Assuming the decisionmaker uses an LG-SSM via this 2TBN structure on batch data over $\mathcal{T}=\{0,1,...,10\}$, we consider an attacker who wishes to maximally disrupt their conditional distributions of the object's true trajectory and speed. The true sensor measurements are given in Table \ref{tab: LG-SSM_trueobs}; the collection rate is known to be $\Delta t =1$. Assuming the object is also known to be moving northeast (i.e., both $z_1$ and $z_2$ tend to increase), the attacker is only willing to corrupt a true observation within an interval centered about $z'_{k,t}$ having a half-width $qz'_{k,t}$ in either direction where $q\in (0,1)$. 

\begin{table}[htbp!]
\centering
\caption{Uncorrupted observations of the object's position in the LG-SSM application} \label{tab: LG-SSM_trueobs}
\resizebox{85mm}{!}{
\begin{tabular}{cccccccccccc} 
\hline
& &\multicolumn{10}{c}{Time ($t$)}\\
& 0 & 1& 2 & 3 & 4 & 5 & 6 & 7 & 8 & 9 & 10\\ \hline
$z'_{1,t}$ & 0.1 & 1.9 & 3.8 & 6.1 &  7.9 & 10.1 & 12.2 & 13.9 & 15.9 & 18.1  & 19.9       \\
$z'_{2,t}$ & 0.2 & 1.1 & 2.3 & 3.1 & 4.2 & 5.1 & 5.9 & 7.1 & 8.2 & 9.4 & 10.2         \\ \hline
\end{tabular}
}
\end{table}

\subsubsection{White-box Setting}


Assume the decisionmaker's true LG-SSM is characterized by the following. The initial distributions are 

\begin{gather*}
Y_{1,0} \sim \mathcal{N}(0,0.01), \quad Y_{2,0} \sim \mathcal{N}(0, 0.01), \\ \dot{Y}_{1,0}\sim \mathcal{N}(2, 0.25), \quad \dot{Y}_{1,0} \sim \mathcal{N}(1, 0.0625).
\end{gather*}

\noindent Since $\Delta t =1$, the transition model when $t\ne0$ is 

\begin{gather*}
    Y_{1,t} \sim \mathcal{N}(Y_{1, t-1} + \dot{Y}_{1, t-1}, 0.01 ), \quad   Y_{2,t}\sim \mathcal{N}(Y_{2, t-1} + \dot{Y}_{2, t-1}, 0.01 ), \\ 
    \dot{Y}_{1,t}\sim \mathcal{N}(\dot{Y}_{1,t}, 0.025), \quad  \dot{Y}_{2,t} \sim \mathcal{N}(\dot{Y}_{2,t}, 0.025),
\end{gather*} 

\noindent and the observation model for all $t \in \mathcal{T}$ is 

\begin{gather*}
    Z_{1,t} \sim \mathcal{N}(Y_{1, t}, 0.04 ), \quad   Z_{2,t} \sim \mathcal{N}(Y_{2, t}, 0.04 ). 
\end{gather*} 

\begin{figure}[htbp!]
    \centering
    \begin{subfigure}[b]{50mm} 
         \centering
         \includegraphics[width=50mm]{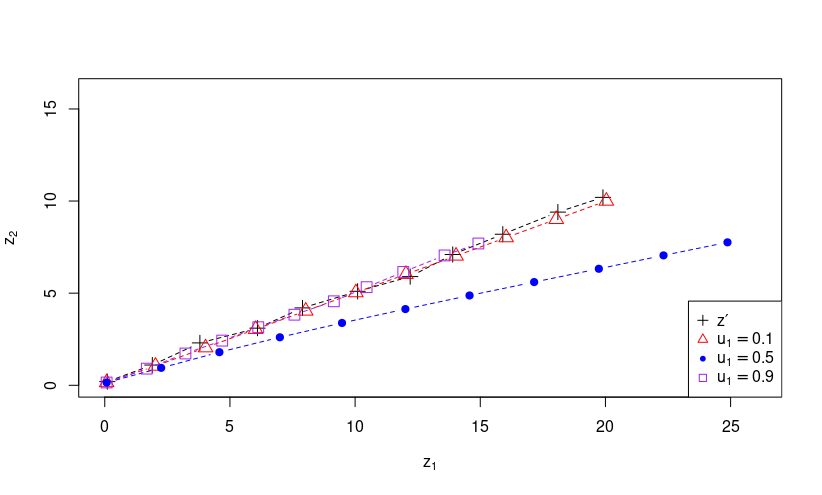}
         \caption{$q=0.25$}
    \end{subfigure}
    \begin{subfigure}[b]{50mm} 
         \centering
         \includegraphics[width=50mm]{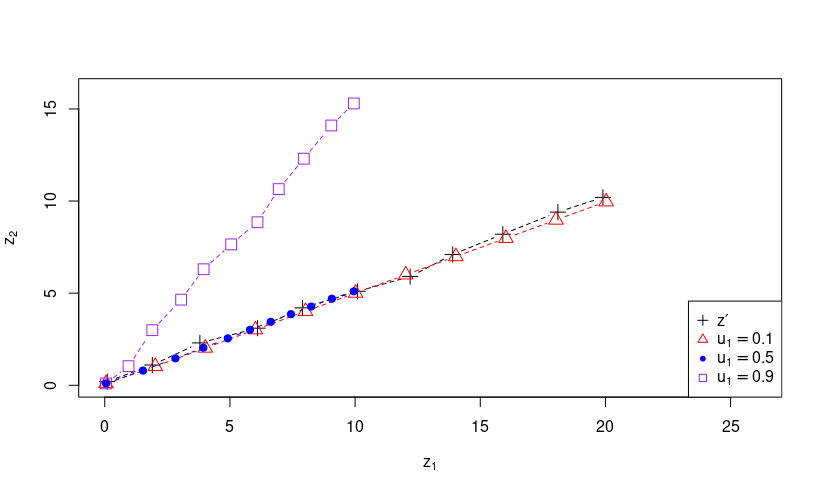}
         \caption{$q=0.5$}
    \end{subfigure}
    \caption{Corrupted observation paths in the white-box LG-SSM Problem}
    \label{fig:LG-SSM_WB}
\end{figure}

\noindent In Figure \ref{fig:LG-SSM_WB}, we plot solutions to the white-box problem identified with CPLEX for varied $q$ and $u$-values against $\mathbf{z}'$. 

\subsubsection{Grey-box Setting} 

\begin{figure}[htbp!]
    \centering
    \begin{subfigure}[b]{50mm} 
         \centering
         \includegraphics[width=50mm]{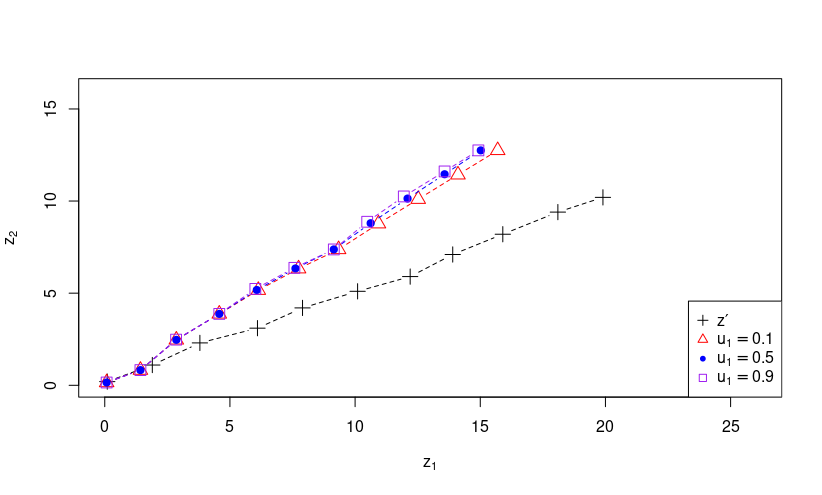}
         \caption{$q=0.25$}
    \end{subfigure}
    \begin{subfigure}[b]{50mm} 
         \centering
         \includegraphics[width=50mm]{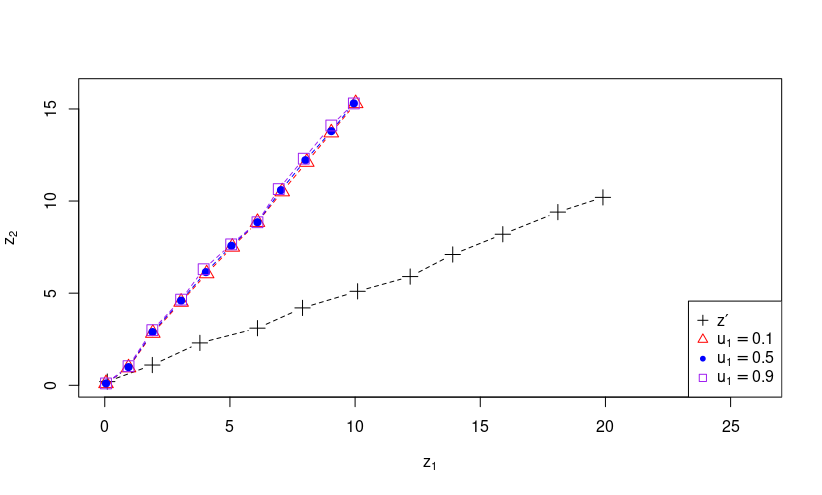}
         \caption{$q=0.5$}
    \end{subfigure}
    \caption{Corrupted observation paths in the grey-box LG-SSM Problem}
    \label{fig:LG-SSM_GB}
\end{figure}

The attacker is uncertain of the decisionmaker's initial distributions, but knows they are univariate normals. The LG-SSM dictates the means of the transition and observation models, but the attacker is uncertain of their variances. The attacker characterizes their beliefs about the unknown means via independent normal distributions centered on the true values having unit variances. We model the unknown variances with an inverse-gamma having a shape parameter equal to two and a scale parameter equal to the true variance. In this manner, the attacker's belief is correct in expectation. In Figure \ref{fig:LG-SSM_GB}, we plot solutions to the grey-box problem using SAA with $J=1000$ for alternative $q$ and $u$-values against $\mathbf{z}'$.

\subsubsection{Implications and Insights}
Whereas our attacks are designed to indiscriminately corrupt inference, herein we focus explicitly on their effects for state estimation. Specifically, we examine patterns in $\mathbf{z}$ and explore their effect on the estimation of $\mathbf{Y}$. Such a task is paramount for, e.g., air traffic control decisions. 


In the white-box setting, we observe that, when $u_1=0.1$, the attacker chooses to select $\mathbf{z}$ relatively close to $\mathbf{z}'$; in so doing, the object's trajectory and velocity coincide more with the true data. However, the behavior is more varied for $u_1=0.5$ and $u_1=0.9$. When $q=0.1$, the corrupted data under $u_1=0.5$ implies a similar velocity but a distinct trajectory. Alternatively, when $u_1=0.9$ for the same $q$-value, the corrupted observations lie upon a similar trajectory but terminate much earlier, thereby implying a lesser velocity. Interestingly, this trading of path and velocity similarity persists when $q=0.5$, but in the opposite direction, That is, the $u_1=0.5$ corrupted data implies a similar trajectory and lesser velocity, whereas $u_1=0.9$ suggest a similar velocity but distinct trajectory. 
This suggest that the attacks balance between obscuring posterior beliefs on the $Y_{1,t}$- and $Y_{2,t}$-values with those about the $\dot{Y}_{1,t}$- and $\dot{Y}_{2,t}$-values. Nevertheless, because our attacks are indiscriminate, the most disruptive (but plausible) $\mathbf{z}$ is chosen as measured by the KL divergence. This behavior is most beneficial when the attacker wish to cause broad versus targeted disruption.

Pivoting to the grey-box setting, for both values of $q$ and for all values of $u_1$, the attacks push the observations further north and less east. The main difference is the degree of corruption; a larger $q$-value induces a larger $\mathcal{Z}$, and the attacks move toward this boundary. The implication is that, either the KL divergence objective remains dominant at $u_1=0.1$, or the two objectives are in less conflict for the given $\mathbf{z}'$ and $\mathcal{Z}$ pairing. This was not the case in the white-box setting. Notably, we observe that the grey-box attacks are qualitatively different than the white-box attacks; the speed and velocity tradeoffs are not apparent at the examined $u_1$-values. This is interesting given that, in expectation, the attacker's beliefs accord with the decisionmaker's model from the previous section. However, echoing the insights from the other applications, we see that such a property does not necessarily imply that the white- and grey-box problems are identical.

This application also highlights that, for practical implementation, the weights the attacker assigns to each objective is informed by the structure of the data and the geometry of $\mathcal{Z}$. For example, suppose it is known that the decisionmaker utilizes a sensor having a resolution of one meter and a quantization error of 0.5 meters. It is therefore conceivable that any corruption within $\pm 1$ meter from the true sensor measurement will not raise suspicion. Forming $\mathcal{Z}$ with a ball of radius one centered at $\mathbf{z}'$, the attacker may reasonably set $u_2=0$ and strictly maximize the KL Divergence. Alternatively, suppose the attacker is willing to risk detection in hopes of further disrupting the conditional inference and forms $\mathcal{Z}$ with a ball of radius two centered at $\mathbf{z}'$. In this case, perturbations beyond the sensors inherent error are more likely to raise suspicions, potentially necessitating a greater $u_2$.

In general, this application exemplifies the threat of data manipulation to state-space models and the autonomous systems that use them (e.g., automated driving systems). This application showcases the vulnerability of LG-SSMs to such attacks and portends the vulnerability of other state-space models used in high-stakes applications as well (e.g., space situational awareness). This is especially relevant given the increased autonomy embedded within military systems \citep{roblin2024}.

\section{Conclusion} \label{sec6}

Recent research in adversarial machine learning has highlighted the effect of adversarial data in computer-vision classification; however, the same opportunities and vulnerabilities exist in alternative settings. Motivated by this observation, we develop tailored attacks against MVGs in both the white- and grey-box settings. In juxtaposition to \citet{kurakin2016adversarial}, we illustrate how a model-specific approach may enable the identification of optimal white-box attacks. 
Moreover, in the grey-box setting, we put into practice principles advanced by \citet{rios2023adversarial} and extended them using techniques from stochastic programming. We demonstrate the efficacy and flexibility of our methods via a collection of applications that attacked an explicit MVG, a linear regression model with Gaussian error and a linear Gaussian state-space model. Within these applications, our attacks were shown to be effective, computationally efficient, and tailorable to varied information conditions. 

Nevertheless, we contend that the study of adversarial statistical modeling is in its infancy and that avenues of future inquiry abound, many directly extending the work herein. For example, whereas we derive Problem WB and GB using the KL-divergence and the log-ratio penalty, these are but one of numerous design choices. If it is more amenable to the attacker, one may replace the KL divergence with an alternative statistical distance. A distinct penalty function may also be leveraged, or it could be omitted and an approach akin to \citet{kurakin2018adversarial} may be adopted wherein only constraints ensure $\mathbf{z}$ is within some $\varepsilon$-neighborhood of $\mathbf{z}'$. Each of these approaches has their relative advantages and disadvantages. For example, leveraging the KL divergence between $\mathcal{P}_{\mathbf{Y| \boldsymbol\mu_{[\mathbf{Z}]}}}$ and $\mathcal{P}_{\mathbf{Y| \mathbf{z}}}$ instead of the log-ratio penalty greatly simplifies analysis, but omits associations in $\mathbf{Z}$ for its penalization. The $\varepsilon$-neighborhood approach is also convenient analytically, but lacks the fine-tuning available in our formulation via its multi-objective optimization construction. Future research may compare and contrast these alternative formulations to discern how their solutions vary systematically.

Additionally, this research considered attacking an MVG; however, by adapting variational methods, our attacks may be used against an arbitrary distribution by approximating it with an MVG and attacking this approximation. Furthermore, our general paradigm toward constructing white- and grey-box attacks against learned models may be extended to other settings. Although the classification setting is relatively well-studied, few authors have yet considered generative, functional or temporal models, among others. Research opportunities exist in these areas for attacks on the model learning process as well as those against a learned model. Finally, given the demonstrated efficacy of our attacks, future research should consider how a decisionmaker may safeguard their MVG inference against an adversary. 

\small
\section*{\small Disclaimer and Acknowledgments}
\noindent The views expressed in this article are those of the authors and do not reflect the official policy or position of the United States Air Force, United States Department of Defense, or United States Government.This research is partially supported by the Air Force Office of Scientific Research (AFOSR) under the Dynamic Data and Information Processing (DDIP) portfolio via project 21RT0867.

\normalsize

\bibliographystyle{model5-names}\biboptions{authoryear}
\bibliography{ref.bib}

\end{document}